\pdfoutput=1

\documentclass[11pt]{article}

\usepackage[preprint]{acl}
\usepackage{hyperref}   
\usepackage{times}
\usepackage{latexsym}
\usepackage{booktabs}
\usepackage[T1]{fontenc}

\usepackage[utf8]{inputenc}

\usepackage{microtype}

\usepackage{inconsolata}

\usepackage{graphicx}
\usepackage[table,dvipsnames]{xcolor}
\usepackage{amsmath,amssymb,amsthm}
\usepackage{float}
\usepackage{multirow}
\usepackage{lineno}
\usepackage{arydshln}
\usepackage{graphicx}
\usepackage{enumitem}
\usepackage{listings}
\usepackage{rotating}
\usepackage{multicol}
\usepackage{amsmath}
\usepackage{cleveref}
\usepackage{bm}
\usepackage{soul}
\usepackage{algorithm}
\usepackage{algpseudocode} 
\usepackage[most]{tcolorbox}
\usepackage{caption}
\usepackage{wrapfig}
\usepackage{subcaption}
\newcommand{\squishlist}{
\begin{list}{{{\small{$\bullet$}}}}
{\setlength{\itemsep}{3pt}      \setlength{\parsep}{1pt}
\setlength{\topsep}{1pt}       \setlength{\partopsep}{0pt}
\setlength{\leftmargin}{1em} \setlength{\labelwidth}{1em}
\setlength{\labelsep}{0.5em} } }
\newcommand{\squishend}{  \end{list}  }

\usepackage{pifont}
\newcommand{\cmark}{\ding{51}} 
\newcommand{\xmark}{\ding{55}} 

\title{Observations and Remedies for Large Language Model Bias \\ in Self-Consuming Performative Loop}

\author{
  Yaxuan Wang$^{1}$,
  Zhongteng Cai$^{2}$,
  Yujia Bao$^{3}$, 
  Xueru Zhang$^{2}$, 
  Yang Liu$^{1}$\\
  $^{1}$University of California, Santa Cruz,
  $^{2}$The Ohio State University, \\
  $^{3}$Center for Advanced AI, Accenture
}

\begin{document}
\maketitle
\begin{abstract}
The rapid advancement of large language models (LLMs) has led to growing interest in using synthetic data to train future models. However, this creates a self-consuming retraining loop, where models are trained on their own outputs and may cause performance drops and induce emerging biases.
In real-world applications, previously deployed LLMs may influence the data they generate, leading to a dynamic system driven by user feedback. For example, if a model continues to underserve users from a group, less query data will be collected from this particular demographic of users. 
In this study, we introduce the concept of \textbf{S}elf-\textbf{C}onsuming \textbf{P}erformative \textbf{L}oop (\textbf{SCPL}) and 
investigate the role of synthetic data in shaping bias during these dynamic iterative training processes under controlled performative feedback.
This controlled setting is motivated by the inaccessibility of real-world user preference data from dynamic production systems, and enables us to isolate and analyze feedback-driven bias evolution in a principled manner.
We focus on two types of loops, including the typical retraining setting and the incremental fine-tuning setting, which is largely underexplored.
Through experiments on three real-world tasks, we find that the performative loop increases preference bias and decreases disparate bias. We design a reward-based rejection sampling strategy to mitigate the bias, moving towards more trustworthy self-improving systems.
\end{abstract}

\section{Introduction}
The widespread integration of Large Language Models (LLMs) into daily applications has raised concerns regarding training on synthetic data~\cite{cai2025stabilizing}. As LLMs become more capable, there are a large amount of generated content that is posted to coding platforms, social media platforms and other platforms on the internet. Such LLM-generated text is getting hard to distinguish from human-generated content~\cite{sadasivan2023can} and might be used to train the next generation of LLMs. 
While human-generated data still exists, access to clean human-authored data is increasingly limited by contamination and cost, prompting widespread reliance on synthetic data.
Consequently, a self-consuming training loop~\cite{shumailov2023curse,  briesch2023large, alemohammad2023self,alemohammad2024self, ferbach2024self} emerges in which future models are trained repeatedly on LLM-generated data from previous generations. 
The recursive training loop on synthetic data may lead to model collapse~\cite{guo2023curious, seddik2024bad, shumailov2024ai,wei2025self}, and bias amplification~\cite{wang2024bias, wyllie2024fairness, chen2024would} or reduction~\cite{chen2024would}.

Unlike static supervised fine-tuning (SFT) with a fixed dataset, online continual learning~\cite{wang2024dealing} collects feedback from humans or AI agents to iteratively train models that are more capable and better aligned.
In socially predictive systems, the model’s performance subsequently influences future data, which is known as performative prediction~\cite{perdomo2020performative, hardt2023performative}. Bias variation in such performative iterative training loops is critical. While~\citet{wyllie2024fairness} demonstrate that unfair feedback loops can lead to a loss of fairness, their analysis does not extend to the data ecosystem of LLMs.

Previous work~\cite{wang2024bias, briesch2023large, kazdan2024collapse} has primarily focused on the self-consuming retraining loop for LLMs, where the next generation model is fine-tuned on the base model using generated synthetic data.
However, in practical scenarios, retraining from the base model may result in the loss of previously acquired knowledge, especially when access to the original training data is limited or restricted.
A more feasible and widely adopted approach in practice is to fine-tune new models based on existing finetuned ones~\citep{gillman2024self}. 
However, this incremental fine-tuning setting in self consuming loop remains underexplored in current research.

\begin{figure}[t]
  \includegraphics[width=\columnwidth]{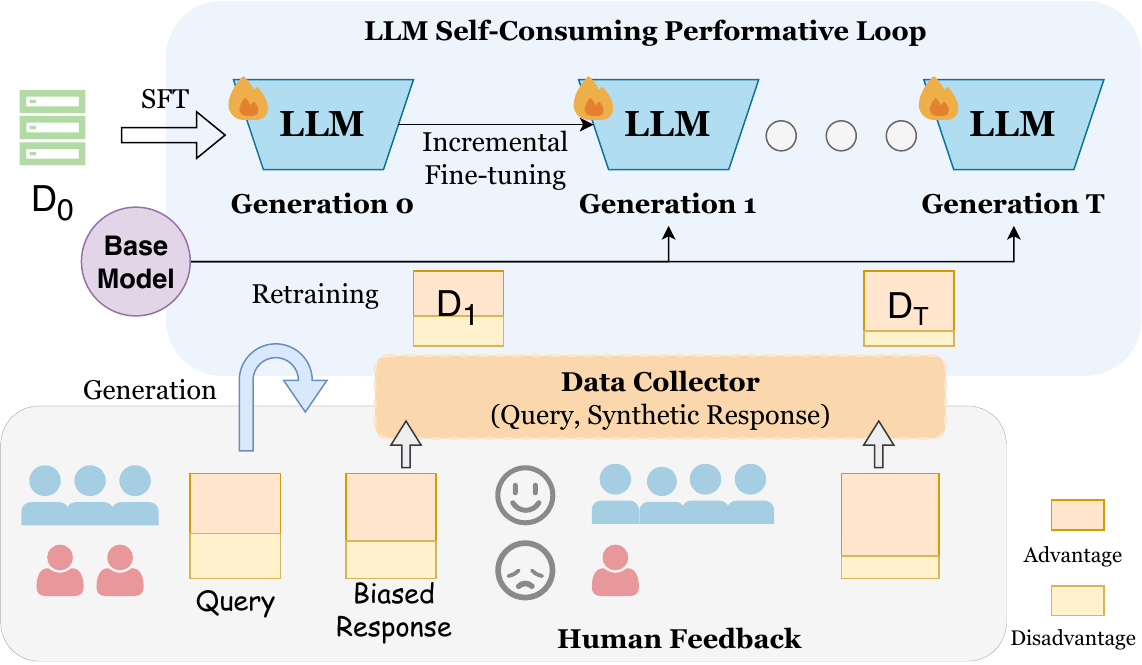}
  \caption{Illustration of the self-consuming performative loop for LLMs. Dynamic human feedback (e.g., an increase in queries from the blue group and a decrease from the pink group) influences both the generation of synthetic data and the subsequent training process.}
  \label{fig:framework}
\end{figure}

In this paper, we introduce the self-consuming performative loop for LLMs (Figure~\ref{fig:framework}), which is based on performative feedback~\cite{perdomo2020performative} and LLM self-consuming training. We analyze bias variation in two settings: a novel incremental fine-tuning loop and the standard retraining loop. We investigate how synthetic data generated by previous models influences bias dynamics during iterative training with controlled human feedback. In particular, we examine how repeated self-consuming training affects both preference bias and disparate performance bias, providing insights that inform safer deployment of future LLMs.

Our findings reveal that iterative fine-tuning/retrain with self-generated data amplifies preference bias and degrades generation quality over time. Interestingly, disparate bias tends to decrease, suggesting a convergence of performance across groups. We also observe that performative feedback accelerates bias amplification in incremental fine-tuning loops,  whereas this phenomenon is less obvious in retraining loops. To address the observed bias amplification, we present a mitigation technique, which integrates reward-guided selection and a reweighting mechanism to control the data sampling process. We also explore several naive rejection sampling~\citep{yuan2023scaling} strategies as baselines. We conduct comprehensive experiments on three real-world tasks, which validate our findings and demonstrate the effectiveness of the proposed mitigation techniques. Our main contributions are as follows:

\squishlist
    \item We introduce the SCPL for LLMs and, for the first time, study the dynamic incremental fine-tuning loop under performative human feedback.
    \item We investigate the impact of synthetic data on bias in the SCPL through extensive experiments and provide several key observations.
    \item We explore various synthetic data curation methods and design a reward-guided selection with reweighting technique to mitigate bias amplification in the performative loop.
\squishend

\section{Related Work}
\textbf{Self-consuming Training Loops.}
Recent work shows that recursively training generative models on their own outputs leads to model collapse, where data quality and diversity degrade over iterations \cite{alemohammad2023self, shumailov2024ai, briesch2023large, cai2025stabilizing}. Theoretically, collapse is inevitable when relying solely on synthetic data \cite{seddik2024bad}, while incorporating fresh real data can stabilize training \cite{bertrand2023stability}. Subsequent analyses provide performance degradation bounds and modified scaling laws \cite{dohmatob2024model, dohmatob2024tale}. Several studies highlight that LLMs are particularly vulnerable in such recursive settings \cite{seddik2024bad, briesch2023large, guo2023curious}, and that data curation implicitly acts as preference optimization in self-training loops \cite{ferbach2024self}.

\textbf{Performative Prediction.}
Performative prediction studies learning systems whose deployment alters the data distribution itself \cite{perdomo2020performative, chen2023performative}. Prior work analyzes stability, convergence, and optimality under these feedback dynamics \cite{hardt2023performative, piliouras2023multi, jin2024addressing}, providing a theoretical lens for self-consuming training loops.

\textbf{Bias in Self-consuming Systems.}
Recent studies show that biases can emerge and amplify in self-consuming training, particularly for LLMs trained on biased human text or synthetic data \cite{wyllie2024fairness, wang2024bias, chen2024would}. Such models may internalize social or political biases present in their training corpora \cite{haller2023opiniongpt, rettenberger2025assessing, wang2024jobfair}. Mitigation strategies have been explored, but remain limited in recursive settings \cite{wang2024bias}.

\textbf{Rejection Sampling.}
Rejection sampling filters synthetic outputs using heuristics or reward models to curate high-quality data for LLM fine-tuning~\cite{yuan2023scaling, toshniwal2024openmathinstruct, khaki2024rs,tong2024dart,li2025fastmcts, pang2025supervised}. Such curation steers models toward high-reward regions~\cite{ferbach2024self}, motivating our rewards-based reweighting approach for bias mitigation in SCPL.

\section{Self-Consuming Performative Loop}

\subsection{Preliminaries}
Formally, let $\mathcal{M}_t$ with parameters $\theta_t$ denote the LLM at generation $t$, and let $T$ be the total number of generations. Let $x$ denote a prompt, $y$ its corresponding response, and $z = (x, y)$ the prompt-response pair, i.e., a single data sample. A dataset $D = [z_1, z_2, \ldots, z_n]$ then consists of $n$ samples.

\textbf{LLM Finetuning Loss }
For a dataset $D$ consisting of $n$ prompt-response pairs $(x, y)$, the fine-tuning loss for a model $\mathcal{M}$ with parameters $\theta$ is defined as:
$\mathbb{L}(D; \theta) = \frac{1}{n} \sum_{(x,y)\in D} \mathcal{L}(x,y;\theta) $

\textbf{Self-Consuming Training Loop }
For LLMs, each training cycle begins with a pre-trained model fine-tuned on recent data via typical supervised fine-tuning~\cite{shumailov2024ai, briesch2023large}. A synthetic dataset $D_t$ is generated using the previous model $\mathcal{M}_{t-1}$, and the next generation model $\mathcal{M}_t$ is trained from scratch (fine-tuned on the base model) either using only generated synthetic data $D_t$ or a mixture of real and synthetic data, depending on the data cycle design (Section~\ref{sec:pre:data-cycle}).

\subsection{Self-Consuming Performative Loop}

We propose a dynamic self-consuming training loop for LLMs (SCPL) where the predictive performance of the previous generation model influences the distribution of generated data, thereby affecting the next generation model. Performative here refers to dynamic, feedback-driven changes in the input distribution.
Under performativity, the next generation model $\mathcal{M}_t$ is fine-tuned on data generated from the previous model $\mathcal{M}_{t-1}$, leading to a performative fine-tuning loss defined as:
\begin{equation}
    \label{eq:finetune}
    \mathbb{L}(D_t(\theta_{t-1}, \mathbb{H}_{t-1}); \theta_t) =  \frac{1}{n} \sum_{(x,y)\in D_t} \mathcal{L}(x,y;\theta_t)
\end{equation}
where $D_t(\theta_{t-1}, \mathbb{H}_{t-1})$ is the dataset induced by $\mathcal{M}_{t-1}$ under real-world human performative feedback $\mathbb{H}_{t-1}$ on a held-out dataset $D_{\text{test}}$ or a training dataset $D_{\text{train}}$, depending on the specific real world application and evaluation objective.

Here, we consider a two-group situation consisting of an advantaged group $D^a$ and a disadvantaged group $D^d$, such that $D = D^a \cup D^d$. We assume an unbiased held out set $D_{\text{test}}$ that contains an equal number of samples from both groups to evaluate the performance. 

If a model performs better on $D^a$, users from this group are more likely to continue interacting and generating data, while users from $D_d$ may disengage, reducing their future representation and exacerbating fairness issues. Our framework models a controlled self-consuming loop in which a single provider  collects, fine-tunes, and repeatedly reuses its own model outputs (e.g. e-commerce recommendation system), enabling systematic study of how internal feedback can amplify or distort demographic preferences. In this study, group-wise sample selection is governed by dynamic functions that simulate practical human preference patterns.

\begin{algorithm}[tb]
   \caption{Self-Consuming Performative Loop}
   \label{alg:self_consuming_loop}
\begin{algorithmic}
   \State {\bfseries Input:} number of generations $T$,
                            data cycle
    \State \textbf{Initialize:} Train base model on real data $D_0$, producing $\mathcal{M}_0$
   \For{$t=1$ {\bfseries to} $t=T$}
        \State Performative sample $D_t$ from $\mathcal{M}_{t-1}$ based on human feedback $\mathbb{H}_{t-1}$
        \If{Accumulation}
        \State $D_t$ $\leftarrow$ $D_t \cup \ldots \cup D_0 $ 
        \EndIf
       \State Train Model $\mathcal{M}_t$ with $D_{t}$ using Eq~\ref{eq:finetune}
   \EndFor
\end{algorithmic}
\end{algorithm}

Algorithm~\ref{alg:self_consuming_loop} presents the SCPL, while Algorithm~\ref{alg:performative_sample} outlines the performative sampling process. 
Moreover, we study two realistic cases for SCPL:

\begin{algorithm}[tb]
   \caption{LLM Performative Sampling}
   \label{alg:performative_sample}
\begin{algorithmic}
   \State {\bfseries Input:} LLM $\mathcal{M}_{t-1}$, $D_{test}$, $D_{\text{candidate}}$, $D_{t-1}$
    \State \textbf{Initialize:} $D_t$: empty list
    \State Calculate performance score $S^a$ on $D_{\text{test}}^a$
    \State  Calculate performance score $S^d$ on $D_{\text{test}}^d$
    \State Human feedback $\mathbb{H} = [S^a, S^b]$
    \State Update the disadvantage ratio $r_d = F(S^a,S^d)$
    \State Select $r_d*|D_{t-1}|$ prompts from $D_{\text{candidate}}^d$
     \State Select $(1-r_d)*|D_{t-1}|$ prompts from $D_{\text{candidate}}^a$

   \For{$x$ {\bfseries in} prompts $X_{\text{candidate, t-1}}^a$}
        \State $y \sim \mathcal{M}_{t-1}(x)$; Add $(x,y)$ to $D_t$
   \EndFor
      \For{$x$ {\bfseries in} prompts $X_{\text{candidate, t-1}}^d$}
        \State $y \sim \mathcal{M}_{t-1}(x)$; Add $(x,y)$ to $D_t$
   \EndFor
   \State {\bfseries Return:} $D_t$
\end{algorithmic}
\end{algorithm}
\textbf{Performative Retraining Loop } 
The next generation model $\mathcal{M}_t$ is retrained from the base model using the synthetic data performative sampling by previous generation model $\mathcal{M}_{t-1}$. This retraining loop is widely study in previous work~\citep{wang2024bias, briesch2023large}.

\textbf{Performative Incremental Fine-tuning Loop }
We consider a more practical scenario reflecting real-world constraints. 
Formally, at generation $t-1$, the model $\mathcal{M}_{t-1}$ is used to generate a new dataset $D_{t}$. This dataset is then used to fine-tune the model $\mathcal{M}_{t-1}$, resulting in the updated model $\mathcal{M}_{t}$.
In user-facing applications with limited computational resources, it is often necessary to incrementally fine-tune the current generation model on top of the previous one, in order to preserve previously acquired knowledge. Retraining from scratch may not be feasible when access to original training data is restricted, for example, due to expired data licenses. In such cases, the current dataset $D_{t-1}$, which only contains newly generated query-response pairs, may no longer suffice to recover past knowledge.
In addition to SFT, we also conduct preliminary experiments using Direct Preference Optimization~\cite{rafailov2023direct}.

\subsection{Data Cycles and Dynamics}
\label{sec:pre:data-cycle}

\textbf{Full Synthetic Data Cycle: }
In the most extreme case, the new model $\mathcal{M}_t$ is only trained on the generated data $D_{t}$ from the last generation~\cite{alemohammad2023self}. Each generation's dataset is of equal size, with $D_0$ representing the initial real human generated dataset. We use this iterative training setup to investigate the most drastic behavioral changes in models under a deployment scenario where the original training data is inaccessible, such as due to expired data licenses or user data deletion requests~\cite{wang2025dragon}. Note that synthetic training data can be obtained by using fresh real prompts collected from users.

\textbf{Expanding Data Cycle (Accumulation): }
This cycle reflects a practical setting in which the existing dataset can be reused during training to improve performance~\citep{briesch2023large}. We accumulate both the original real data $D_0$ and the generated synthetic data across generations to form the training dataset. Specifically, the accumulated dataset at generation $t$ is $\mathcal D_{t} = D_0 \cup D_1 \cup \ldots \cup D_t$. Prior work has shown that such accumulation can slow down model collapse~\cite{kazdan2024collapse} and help mitigate bias~\cite{wang2024bias}.

\textbf{Dynamics: } 
Dynamics refer to the changing proportion between the advantaged and disadvantaged groups, which also indicate the performative impact. We primarily study a \textbf{controlled linear dynamic setting}, where the group ratio is explicitly manipulated in experiments. Starting from an initial disadvantaged group ratio in $D_0$, the proportion decreases linearly over generations to simulate the effect of human feedback $\mathbb{H}$ under controlled conditions. We also evaluate a \textbf{fixed-ratio setting} as a baseline, where the disadvantaged ratio remains constant while using fresh real prompts. 
The scenario in which previous prompts are repeatedly used to generate synthetic data, and the group ratio remains unchanged across generations, is referred to as \textbf{Non-dynamic}, which is the typical self-consuming training loop~\cite{wang2024bias,briesch2023large,alemohammad2023self}.

\section{Measuring the Bias}
\subsection{Overview}
We focus on measuring bias variation in SCPL through extensive experiments on three practical tasks: news continuation, creative preference dissection, and math problem solving. The first two evaluate preference bias, reflecting the model’s inclination toward the advantaged group~\cite{liu2023trustworthy}. The third assesses disparate performance~\cite{liu2023trustworthy}, where the model's accuracy differs between groups (easy versus hard math questions). We observe that: 
(1) Synthetic data in SCPL tends to amplify preference bias while reducing disparate bias.
(2) Performative dynamics accelerate preference bias amplification compared to non-dynamic self-consuming loops in both fine-tuning and retraining settings, though the effect depends on the task and training regime.
(3) Accumulation can slow the increase of preference bias and mitigate the decline in generation quality. It can alleviate the reduction in disparate bias and help preserve math problem-solving ability.

\subsection{Reward-based Reweighting Sampling}
To reduce the observed preference bias amplification in SCPL, we propose a bias mitigation technique based on rejection sampling to enhance fairness in the SCPL. Specifically, for a prompt $x$ from $D_{candidate}$, we generate $k$ independent responses $\bar{y}_1, \ldots, \bar{y}_k \sim \mathcal{M}_{t-1}(x)$. A reward function $R(x, \bar{y})$ is then used to select the highest scoring sample or one that meets predefined criteria. 
We propose a modular and extensible reward-based reweighting method to reduce bias while maintaining generation quality. 
Formally, let $\mathcal{R} = \{r_1, r_2, r_3\}$ denote a set of predefined rules, where $r_1(x, \bar{y}) \in [0,1]$ represents a continuous score measuring the generation quality, and $r_2(x, \bar{y})$ indicates whether the output $\bar{y}$, have the same preference as the input $x$. $r_3$ refers to any possible task-relevant extension rules.
Each rule is associated with an importance weight, denoted by $\alpha_1$ and $\alpha_2$, respectively. The reward function is then defined as:
\begin{equation}
\label{eq:reward}
    R(x,\bar{y}) = \alpha_1 \cdot r_1(x, \bar{y}) + \alpha_2 \cdot  r_2(x,\bar{y}) + r_3
\end{equation}
We integrate reward-guided sample selection with a reweighting scheme, enabling dynamic control over the sampling process to mitigate bias. The idea is to generate and select more data from the disadvantaged group by adjusting the selection criteria and increasing the sample size $k$. (Algorithm~\ref{alg:mitigation})

Regarding rule extension, our framework is designed to be plug-and-play: one can incorporate additional rules that capture properties such as sentiment, stylistic tone, format constraints, or even a trained reward model that returns a score depending on downstream goals. Each rule outputs a reward signal, and these are combined linearly (with tunable weights) to produce the final reward score used in sampling and training. 
When only a single trained reward model is used with its weight set to 1, this formulation reduces to the standard case of using a single, well-trained or a zero-shot reward model. Note that our mitigation strategy does not rely on access to ground-truth labels used in bias evaluation; instead, we use proxy metrics to estimate the degree of bias present in the model.

\section{Experiments}
We consider several different settings: (1) \textbf{self-consuming training loop} (Non-dynamic): We compare synthetic data (\textit{syn}) with real data (\textit{real}). In the real setting, all synthetic data is replaced with real data to examine how performance changes.
(2) \textbf{SCPL }(Dynamic): We evaluate two variants: a controlled linear dynamic version (\textit{syn-dynamic}) and a fixed-ratio version (\textit{syn-dynamic-fr}). As a baseline, we also replace the synthetic data in the controlled linear dynamic setting with real data (\textit{real-dynamic}).
(3) \textbf{Accumulation-enhanced performative loop}: We further apply data accumulation to the proposed SCPL (\textit{syn-dynamic-accu}). We also apply accumulation to real data under the same dynamic control (\textit{real-dynamic-accu}). 
Our experiments involve Qwen2.5-1.5B, Qwen2.5-Math-1.5B, Llama2-7B, Qwen2.5-7B, Qwen2.5-Math-7B, and Llama3.1-8B.
More experimental details and results, including self-consuming DPO and beyond self-consuming analyses, are provided in Appendix~\ref{app:exp}, ~\ref{app:sec-dpo}, \ref{app:sec-beyond-self}.

\subsection{News Continuation Task}
\label{exp:news}
This task examines the political bias of LLMs within the US political spectrum by evaluating their performance on news continuation~\citep{maslej2023artificial}. 
Following~\citet{wang2024bias}, we use 6458 news articles from the Webis-Bias-Flipper-18 dataset~\citep{chen2018learning}, selecting left- and right-leaning articles to construct 256-token prompt–completion pairs. At the initial iteration, the real dataset contains 5,000 samples (3,000 left-leaning and 2,000 right-leaning), where right-leaning articles are treated as the advantaged group and left-leaning articles as the disadvantaged group for preference bias analysis. We evaluate model performance on a held-out unbiased dataset and additionally report accuracy results on the standard MMLU benchmark~\cite{hendryckstest2021}.

\textbf{Evaluation Metric }
(1) \textbf{Right lean bias (Bias Score)} is the fraction of generated continuations classified as right-leaning on the unbiased test dataset with 1000 articles. For each article, the model deterministically generates the next 256 tokens, and a pretrained classifier following~\cite{wang2024bias} predicts the political leaning of the generated continuation (right-lean or left-lean). An unbiased model is expected to yield a score close to 0.5. 
(2) \textbf{Generation Quality (GQ)} is the average score for all generated articles using the Gibberish Detector\footnote{Please refer to \href{https://huggingface.co/madhurjindal/autonlp-Gibberish-Detector-492513457}{https://huggingface.co/madhurjindal/autonlp-Gibberish-Detector-492513457}}. The detector identifies the incoherent of nonsensical text and has four Gibberish score: 3 for clean, 2 for mild gibberish, 1 and 0 for noise.

\begin{figure*}[h]
    \centering
    \begin{minipage}[b]{0.63\linewidth}
        \centering
        \includegraphics[width=\linewidth]{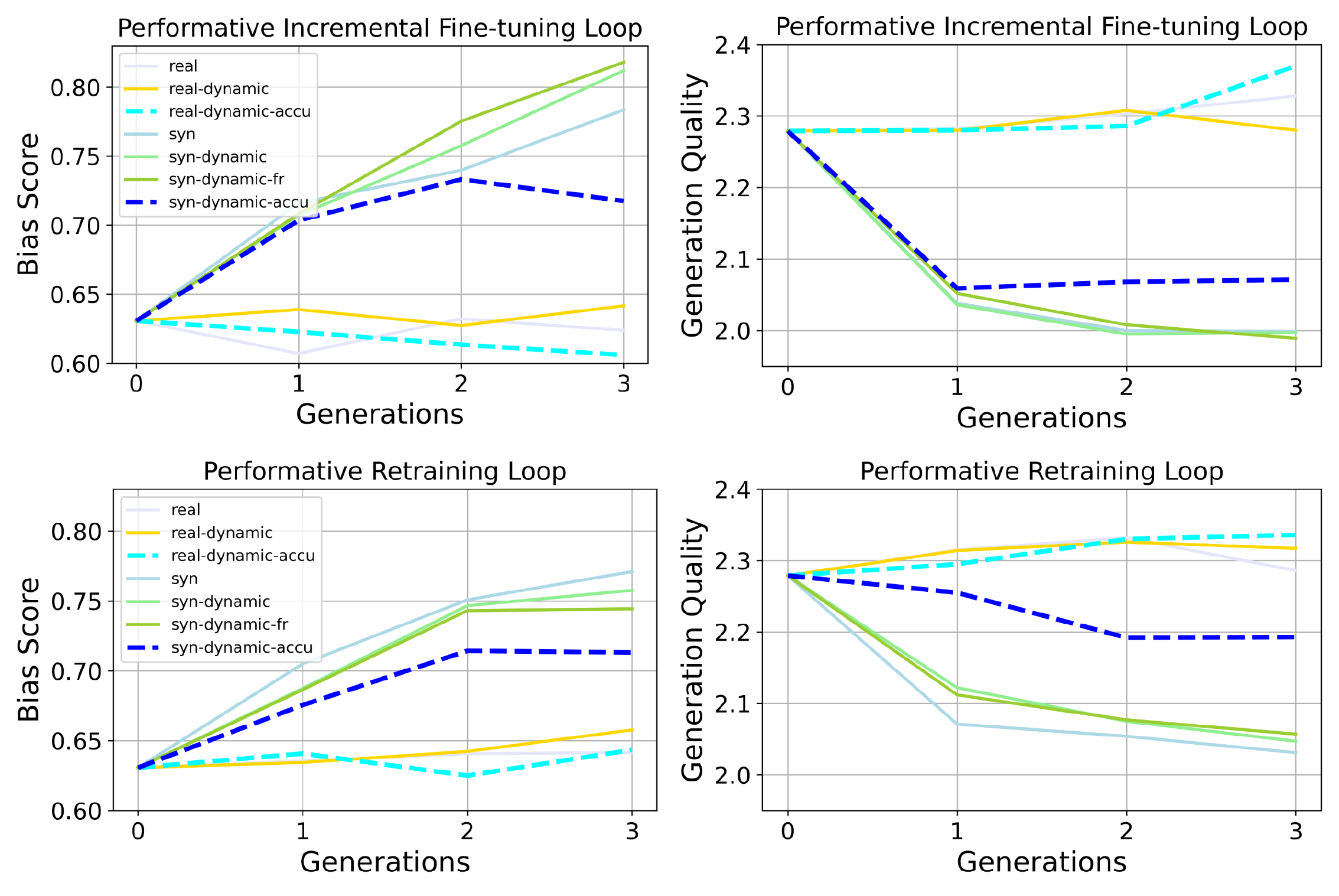}
        \subcaption{News Continuation}
        \label{fig:news}
    \end{minipage}
    \begin{minipage}[b]{0.35\linewidth}
        \centering
        \includegraphics[width=\linewidth]{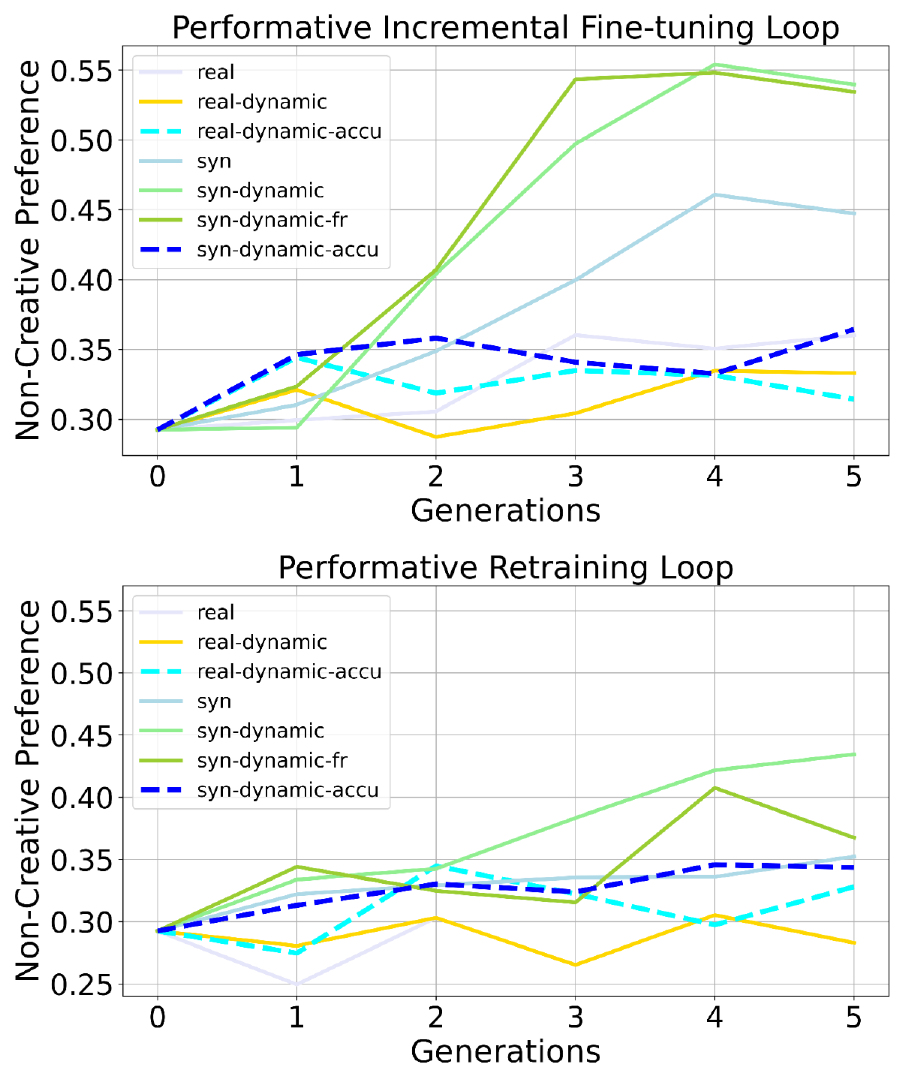}
        \subcaption{Preference Dissection}
        \label{fig:preference}
    \end{minipage}
    \vspace{-3mm}
    \caption{The preference bias and generation quality on two tasks using Qwen2.5-1.5B.}
\end{figure*}

\begin{table}[t]
\centering
\label{tab:performative_incremental}
\resizebox{\linewidth}{!}{
\begin{tabular}{lcccc}
\hline
 & $\textbf{$\mathcal{M}_0$}$ & $\textbf{$\mathcal{M}_1$}$ & $\textbf{$\mathcal{M}_2$}$ & $\textbf{$\mathcal{M}_3$}$  \\
\hline
\multicolumn{5}{c}{\textbf{Preference Bias}} \\
\hline
Real-dynamic & 0.5587 & 0.5727 & 0.5983 & 0.5980 \\
Syn-dynamic  & 0.5587 & 0.6327 & 0.6813 & 0.7136 \\
\hline
\multicolumn{5}{c}{\textbf{Generation Quality}} \\
\hline
Real-dynamic & 2.538  & 2.560  & 2.504  & 2.581 \\
Syn-dynamic  & 2.538  & 2.284  & 2.250  & 2.252 \\
\hline
\multicolumn{5}{c}{\textbf{MMLU}} \\
\hline
Real-dynamic & 0.5820 & 0.5786 & 0.5833 & 0.5842 \\
Syn-dynamic  & 0.5820  & 0.5728  & 0.5661  & 0.5652 \\
\hline
\end{tabular}
}
\caption{Results in the performative incremental fine-tuning loop using Llama3.1-8B on News task.}
\label{app:tab-llama3-finetuning}
\end{table}

\begin{table}[t]
\centering
\label{tab:performative_incremental}
\resizebox{\linewidth}{!}{
\begin{tabular}{lcccc}
\hline
 & $\textbf{$\mathcal{M}_0$}$ & $\textbf{$\mathcal{M}_1$}$ & $\textbf{$\mathcal{M}_2$}$ & $\textbf{$\mathcal{M}_3$}$  \\
\hline
\multicolumn{5}{c}{\textbf{Preference Bias}} \\
\hline
Real-dynamic & 0.5587 & 0.5638 & 0.5695 & 0.6090 \\
Syn-dynamic  & 0.5587&	0.6219 & 0.6478	& 0.6485 \\
\hline
\multicolumn{5}{c}{\textbf{Generation Quality}} \\
\hline
Real-dynamic & 2.538	&2.564 &2.544 &2.450 \\
Syn-dynamic  & 2.538&2.314 &2.303&2.340 \\
\hline
\multicolumn{5}{c}{\textbf{MMLU}} \\
\hline
Real-dynamic & 0.5820 & 0.5773 &  0.5812 & 0.5838\\
Syn-dynamic  & 0.5820 & 0.5803 & 0.5622 & 0.5662 \\
\hline
\end{tabular}
}
\caption{Results in the performative retraining loop using Llama3.1-8B on News task. }
\label{app:tab-llama3-retraining}
\end{table}

\textbf{Results } Preference bias increases more rapidly in the SCPL than in the Non-dynamic loop under incremental fine-tuning. This is likely because bias in the dynamic case can stem from both the increasingly biased fine-tuned model and the performatively generated synthetic data. 
However, under the retraining setting, it shows a slower increase and only marginal differences compared to the non-dynamic loop.
As shown in Figure~\ref{fig:news}, using real data in the iterative training process results in a relatively stable bias trajectory with minimal variation. 
The generation quality decreases at a similar rate in retraining and incremental finetuning settings.

\begin{figure}[t]
    \centering
    \begin{minipage}[b]{0.49\linewidth}
        \centering
        \includegraphics[width=\linewidth]{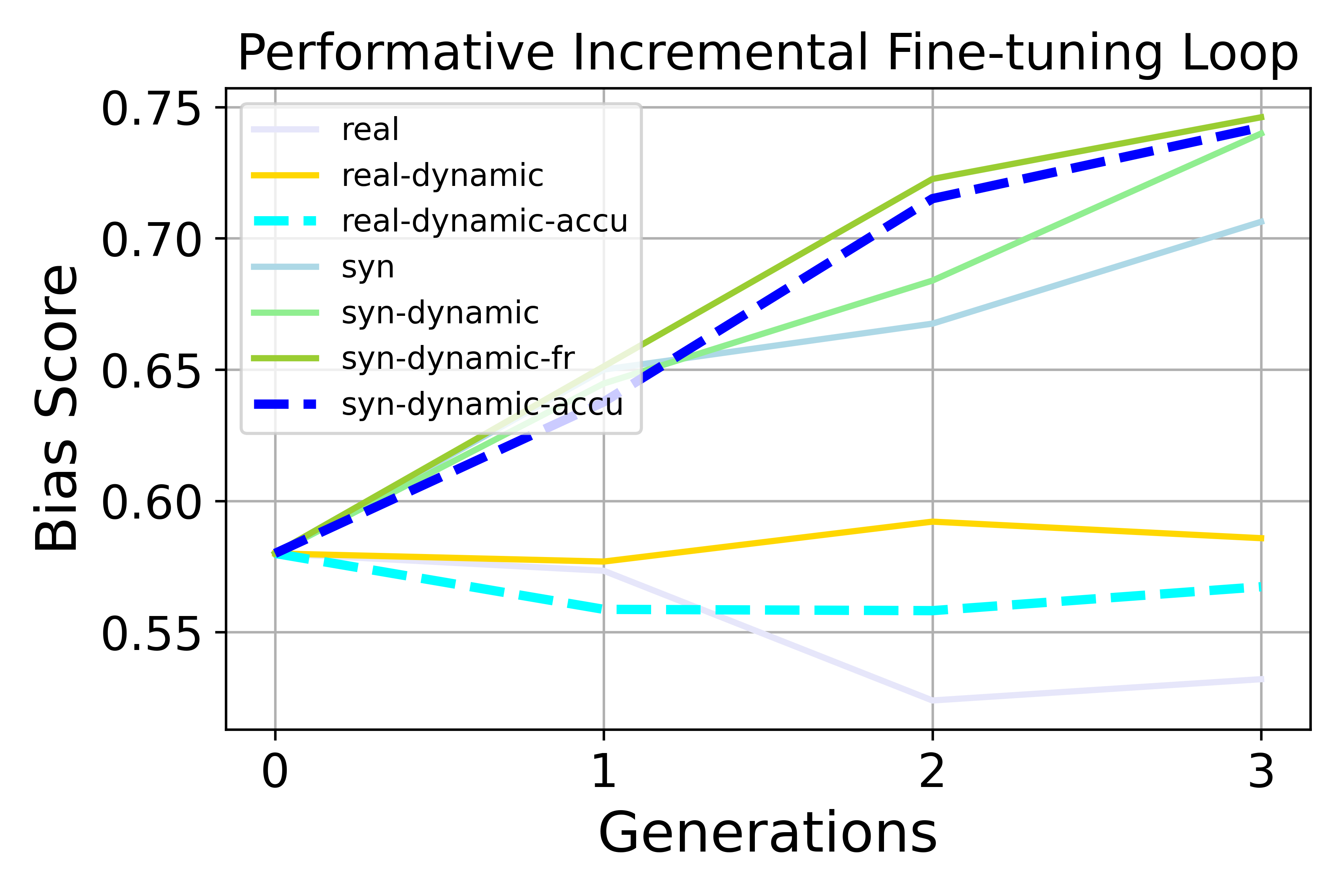}
    \end{minipage}
    \hfill
    \begin{minipage}[b]{0.49\linewidth}
        \centering
        \includegraphics[width=\linewidth]{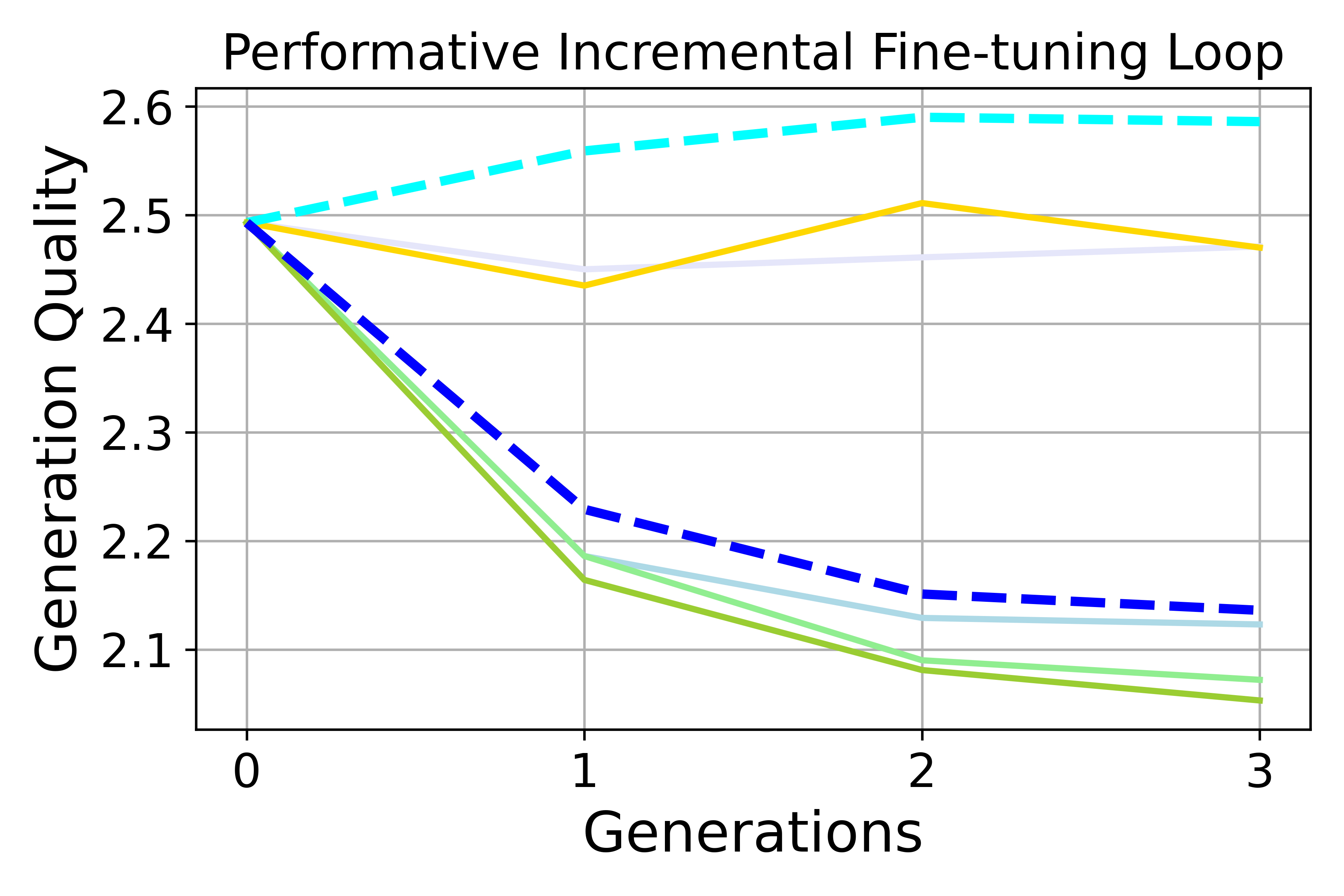}
    \end{minipage}
    \caption{The preference bias and generation quality on news task using Llama2-7B in incremental fine-tuning loop.}
    \label{fig:news-llama2}
\end{figure}

Tables~\ref{app:tab-llama3-finetuning} and~\ref{app:tab-llama3-retraining} show that the observed bias trends are consistent across model families. For completeness, we also report results on the MMLU benchmark. Accuracy remains stable at approximately 0.58 in the real-dynamic setting. In contrast, under the syn-dynamic setting, accuracy decreases at a similar rate for both incremental fine-tuning and retraining, with modest reductions of about 0.02 and 0.05, respectively.
As our primary goal is to analyze the SCPL, standard benchmarks are less sensitive to group-specific distributional shifts and therefore may not fully capture performative bias dynamics.
In Figure~\ref{fig:news-llama2}, the bias score increases more rapidly in dynamic setting, and the GQ declines faster under the incremental fine-tuning loop.

Accumulation helps mitigate the amplification of preference bias and the degradation of generation quality.
However, it does not reduce preference bias compared to the original bias or achieve performance comparable to the real-data setting; it merely slows the rate at which bias increases.
Despite this partial mitigation, we do not observe that accumulation reduces bias toward the neutral line (0.5), as stated in~\citet{wang2024bias}. Therefore, while accumulation can serve as a mitigation strategy for both generation quality and preference bias, its effect on bias remains limited and we need more efforts on mitigation strategies.

\begin{figure*}
    \centering
    \begin{minipage}[b]{0.32\linewidth}
        \centering
         \includegraphics[width=\columnwidth]{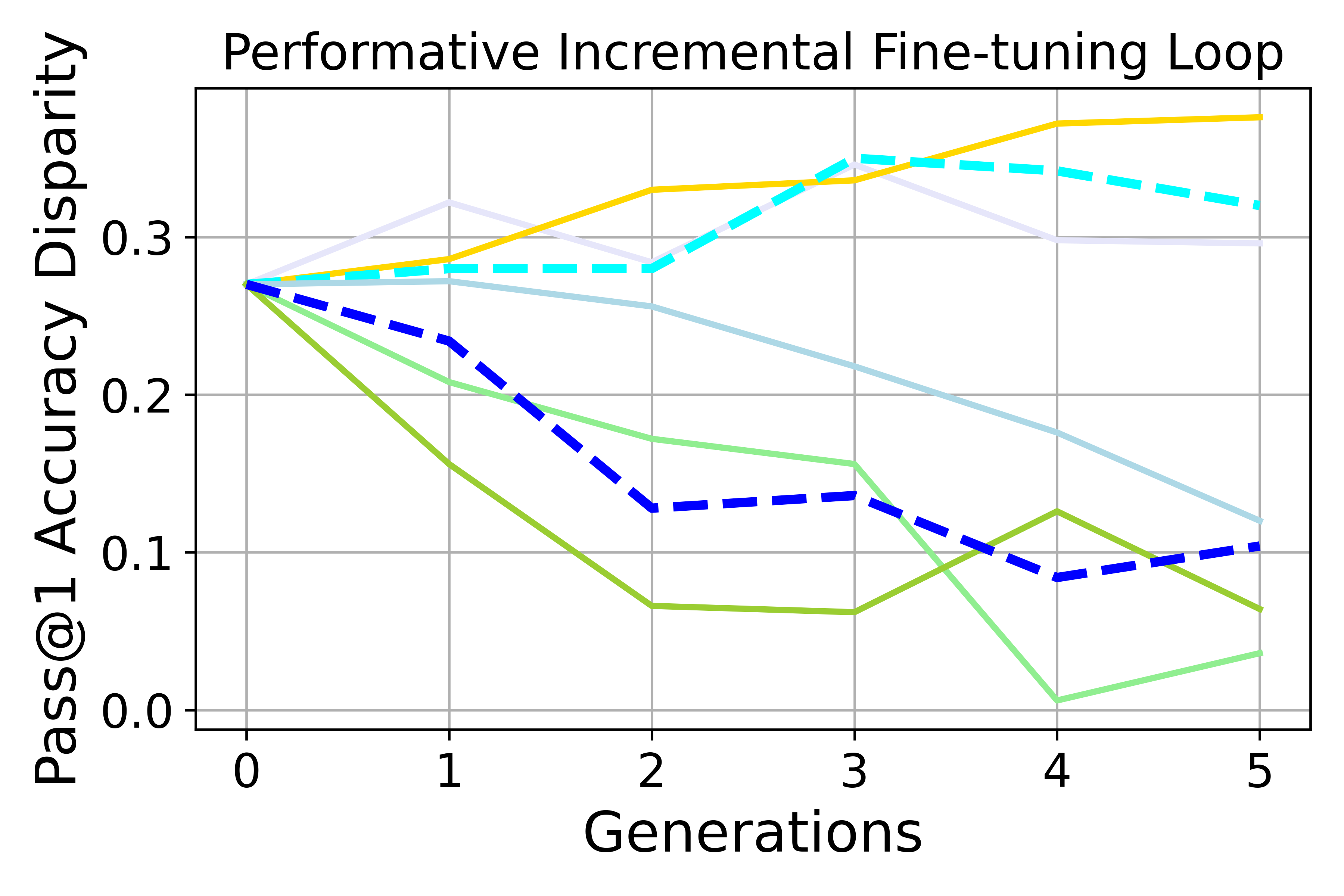}

    \end{minipage}
    \begin{minipage}[b]{0.32\linewidth}
        \centering
         \includegraphics[width=\columnwidth]{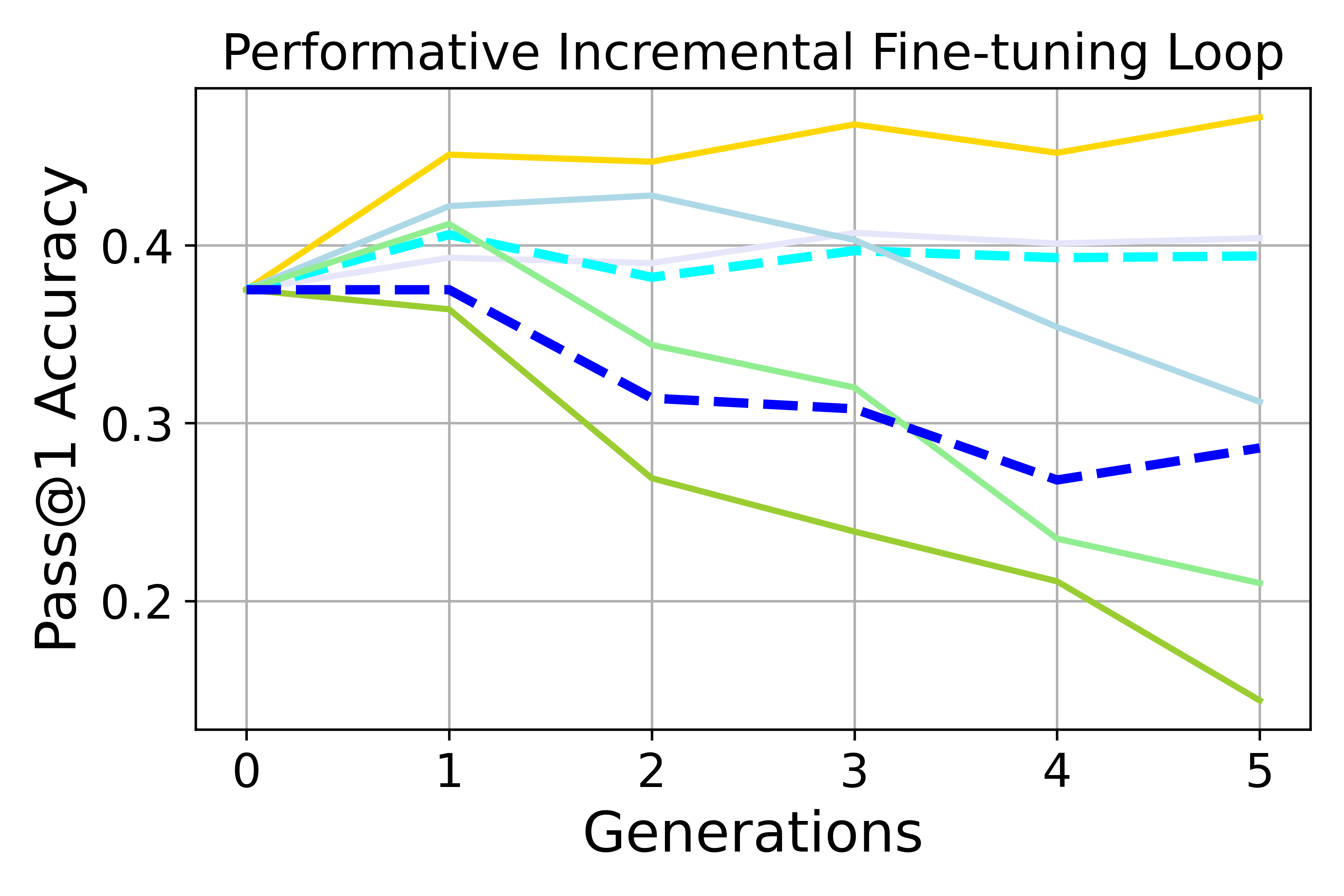}
    \end{minipage}
        \begin{minipage}[b]{0.32\linewidth}
        \centering
         \includegraphics[width=\columnwidth]{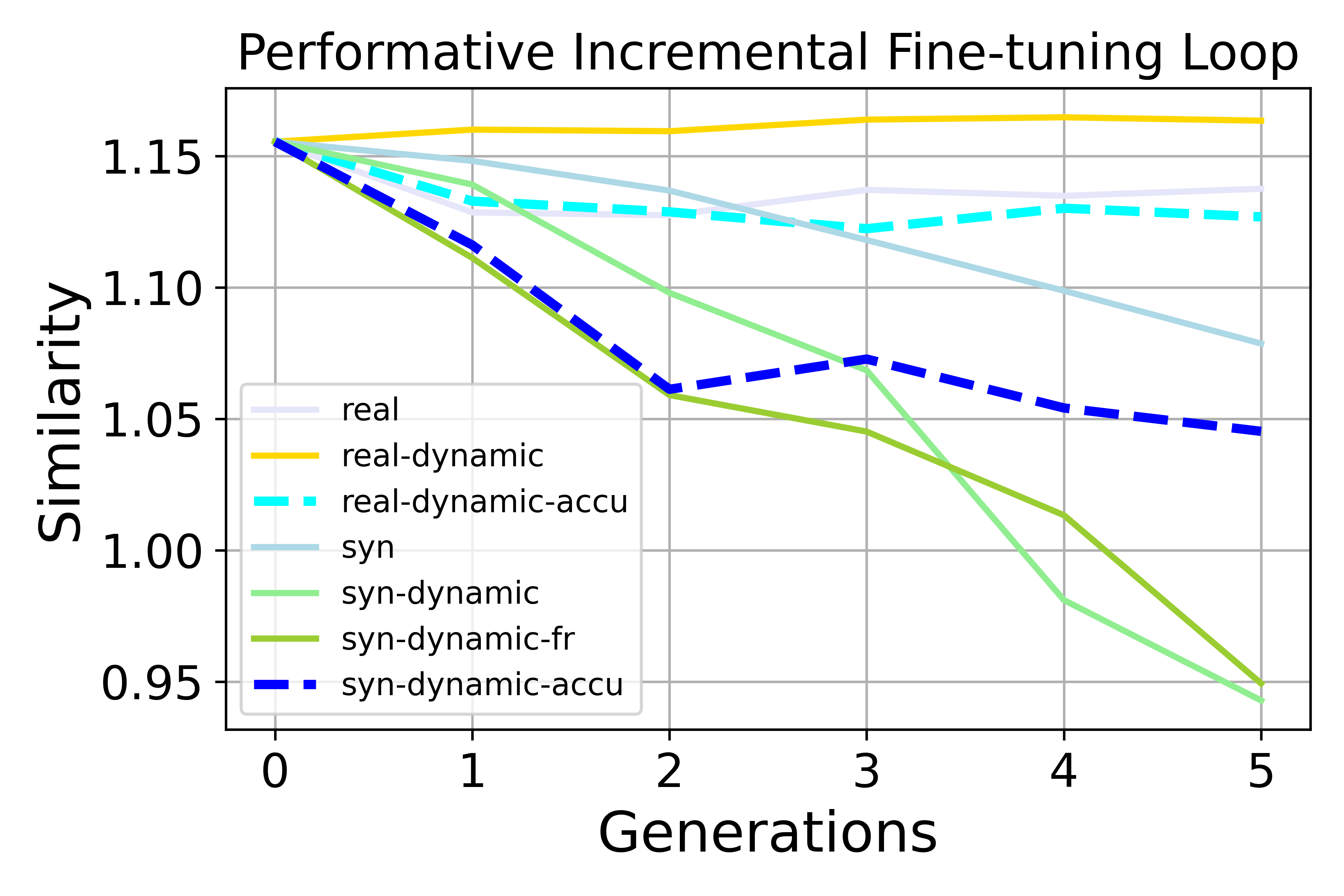}
    \end{minipage}
  \caption{The disparate bias and mathematical problem solving ability on Math task using using Qwen2.5-Math-7B. }
  \label{fig:math-finetuning}
\end{figure*}

\begin{figure*}[h]
    \centering

    \includegraphics[width=\linewidth]{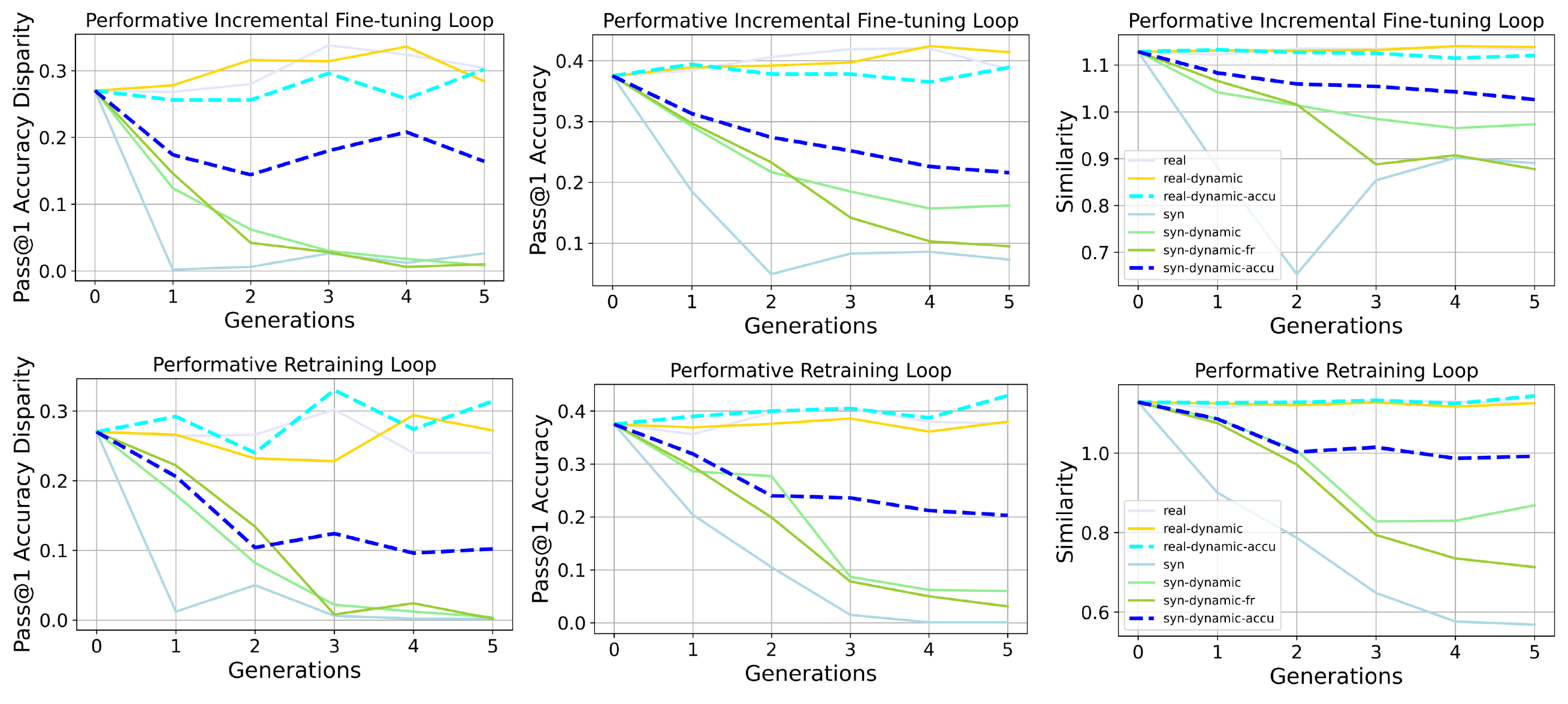}
    
    \vspace{-3mm}
    \caption{The disparate bias and mathematical problem solving ability on Math task using Qwen2.5-Math-1.5B.}
    \label{fig:numina}
\end{figure*}

\subsection{Preference Dissection Task}
\label{exp:preference}
In this task, we explore LLMs’ preference toward non-creative attributes.
We use the knowledge-awre dataset Dolly~\cite{DatabricksBlog2023DollyV2} as the advantage group with 15,000 samples and creative writing dataset ShareGPT~\footnote{\href{https://huggingface.co/datasets/Nitral-AI/Creative_Writing-ShareGPT}{Huggingface: Nitral-AI/Creative\_Writing-ShareGPT}} as the disadvantage group with 6,650 samples. 
We adopt the Preference Dissection dataset~\cite{li2024dissecting} as test dataset.

\textbf{Evaluation Metric } We compute the average preference probability for the non-creative attribute across all ten scenarios in the test dataset~\cite{li2024dissecting} as the preference bias. Since the raw preference scores consistently fall within the range $[0.4,0.6]$ (0.5 denotes neutral), we apply min-max normalization to rescale them. The normalized values are denoted as \textbf{Non-Creative Preference}.

\textbf{Results } Non-Creative Preference Bias increases faster in the SCPL under both incremental fine-tuning and retraining settings.
In the Non-dynamic self-consuming setting, the change in bias follows a trend similar to that of the accumulation-enhanced loop. In contrast, the bias increases more obviously in the self-consuming performative incremental fine-tuning loop, as shown in Figure~\ref{fig:preference}.
Under both fine-tuning and retraining settings, using accumulation effectively mitigates bias and achieves performance close to that of the real-data setting.

\subsection{Math Problem Solving Task}

\begin{figure*}[h]
    \centering
    \begin{minipage}[b]{0.65\linewidth}
        \centering
        \includegraphics[width=\linewidth]{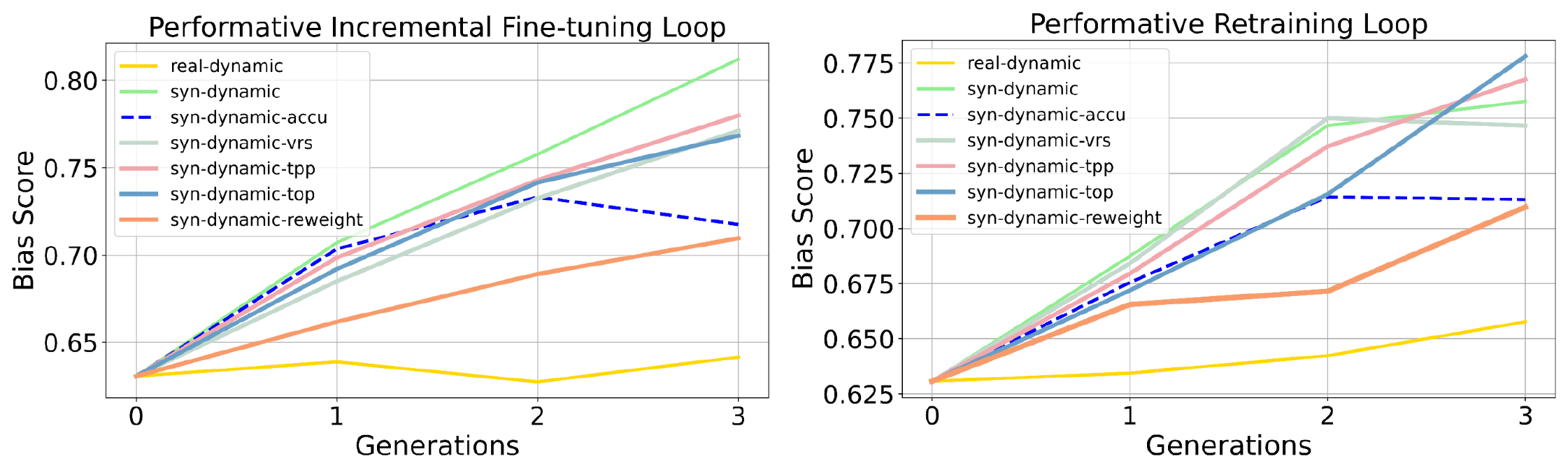}
        \subcaption{News Continuation}
        \label{fig:news-mitigation}
    \end{minipage}
    \begin{minipage}[b]{0.34\linewidth}
        \centering
        \includegraphics[width=\linewidth]{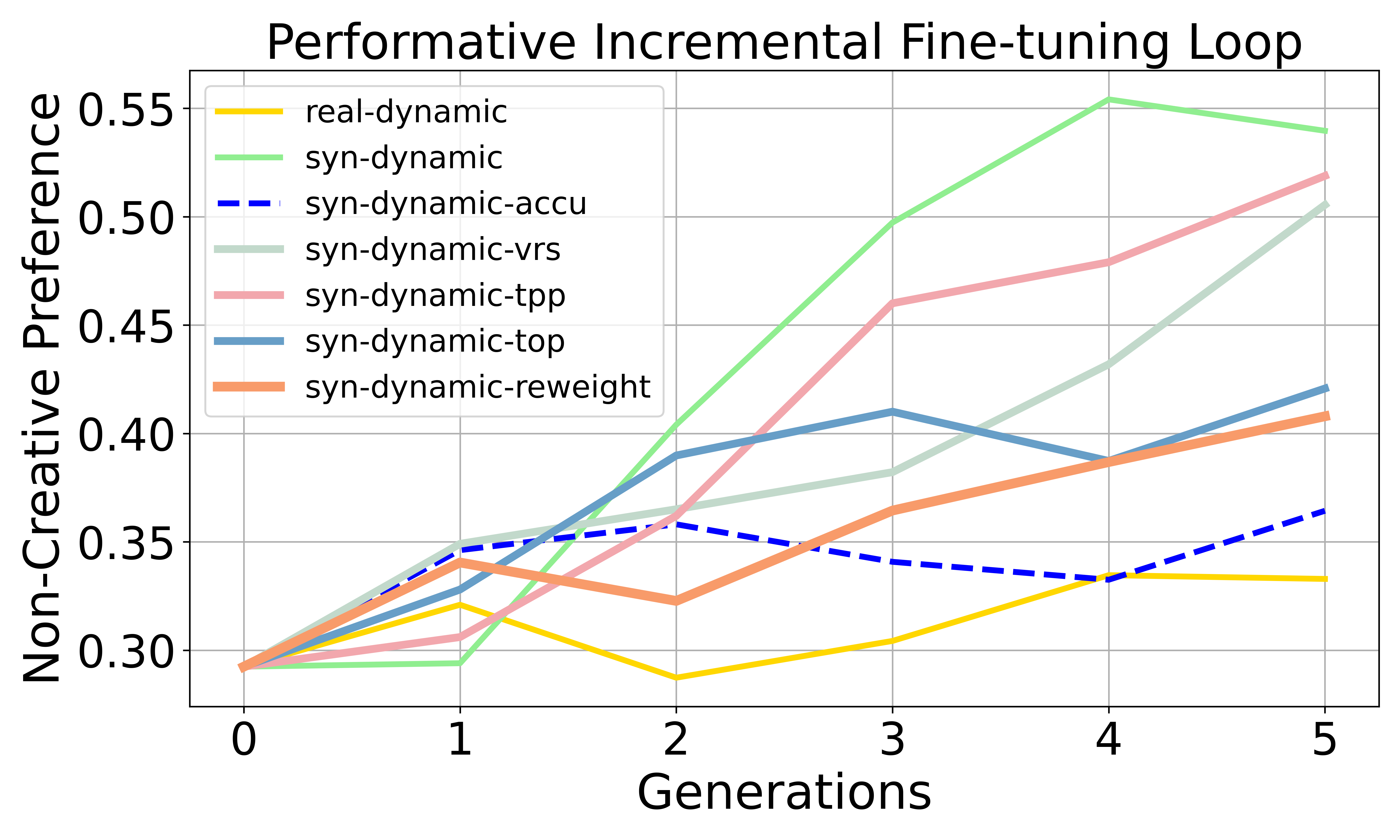}
        \subcaption{Preference Dissection}
        \label{fig:preference-mitigation}
    \end{minipage}
    \vspace{-6mm}
    \caption{The performance of different mitigation methods on two tasks using Qwen2.5-1.5B.}
\end{figure*}

In this task, we use the NuminaMath dataset~\citep{li2024numinamath} and divide the problems into easy and hard groups. Easy problems are considered the advantage group, as they more commonly appear in real-world usage. Over time, the model’s math-solving ability may become skewed due to repeated exposure to easy problems, potentially degrading its performance on harder ones~\citep{tong2024dart}. Our goal is to examine how LLMs perform when exposed to different proportions of easy and hard problems in SCPL. The performance gap between these two groups is treated as disparate bias.

\textbf{Evaluation Metric }
(1) \textbf{Disparate Bias} is the disparity of Pass@1 Accuracy for both groups. Pass@1 Accuracy is the accuracy of solving problems correctly. (2) \textbf{Similarity} is the similarity between the generated response and the ground truth answers. Specifically, it utilizes the sum of ROUGE-L~\cite{lin2004rouge} and Bertscore~\cite{zhang2019bertscore} to represent the problem solving ability.

\textbf{Results }
Disparate bias decreases slower in the SCPL, along with the decrease of overall accuracy and similarity score, as shown in Figure~\ref{fig:math-finetuning} and~\ref{fig:numina}.
This behavior is partly explained by the fact that math performance relies heavily on previously learned knowledge, while synthetic data tends to be lower quality, leading to reduced group disparities but overall performance degradation. In addition, human preferences often favor easier questions, resulting in fewer hard examples over iterations; since LLMs typically struggle more on difficult problems, this shift slows the apparent performance drop as the data distribution becomes simpler.
We further observe that the linear dynamic setting better preserves math-solving ability compared to the size-fixed dynamic. The trend in disparate bias reduction is similar across both settings. 
Notably, model scale plays a critical role: for smaller models such as Qwen2.5-Math-1.5B, performative dynamics slow both performance and bias degradation, whereas for larger models like Qwen2.5-Math-7B, SCPL leads to faster performance decline compared to non-dynamic training (Figure~\ref{fig:math-finetuning}). This suggests that larger models are more sensitive to distributional shifts in synthetic data, and dynamic sampling may disrupt their learned balance by reducing exposure to harder or more diverse examples. In SCPL, Accumulation helps reduce the decline in disparity and math-solving ability.

\subsection{Bias Mitigation Strategy}

We evaluate the effectiveness of our proposed bias mitigation method, which curates less biased data using a predefined rule-based reward function (Eq.~\ref{eq:reward}) and a reweighting technique.

\textbf{Setup } We set $k = 4$, $\alpha_1 = 1.0$, and $\alpha_2 = 3.0$. 
In News task, $r_1$ evaluates the fluency of generated responses using a Gibberish Detector, which assigns a score based on generation quality. Since topic consistency is crucial in this task, we use content similarity as a proxy to assess whether the preference between the input and output remains aligned. $r_2$ calculates the cosine similarity between the generated content and the ground-truth continuation. If the similarity exceeds a predefined threshold, the rule returns +1; otherwise, it returns -1. The same for Preference Dissection task.

\textbf{Baseline } We evaluate three naive rejection sampling methods: Vanilla Rejection Sampling (VRS): Randomly selects one from responses that meet criterias following~\citet{li2025fastmcts}.
Top per Prompt (TPP): Selects the response with the highest reward for each prompt individually.
Top Overall Prompt (TOP): Selects the top $n$ responses with the highest rewards across the entire dataset. We also add Accumulation~\cite{wang2024bias} as one baseline.

\textbf{Results }
The proposed reward-based reweighting sampling achieves the best overall performance for bias mitigation.
Among the five strategies evaluated, the reweight variant demonstrates the most effective bias mitigation on the News task under both retraining and fine-tuning settings, as shown in Figure~\ref{fig:news-mitigation}.
On the Preference Dissection task, our method ranks second in performance, as the accumulation-based approach achieves results close to those obtained using real data (Figure~\ref{fig:preference-mitigation}).

\section{Conclusion}
In this work, we introduce the SCPL for LLMs and study how synthetic data impacts bias under iterative training with performative feedback. Focusing on the underexplored incremental fine-tuning setting and typical retraining setting, we show that preference bias increases in both settings. We also observe the decline of disparate bias.
To address the bias amplification, we propose a reward-based sampling method. Extensive experiments on three real world tasks demonstrate our findings and the effectiveness of the mitigation method.
Further analysis of self-consuming reinforcement learning loops and settings beyond self-consumption remains an important direction for future work.

\newpage

\section*{Acknowledgements}
Y. Wang and Y. Liu are partially supported by the National Science Foundation (NSF) under grants IIS-2143895 and IIS-2416896. 
Z. Cai and X. Zhang are funded in part by the National Science Foundation under award IIS-2202699 and IIS-2416895.

\section*{Limitations}

One limitation of our approach lies in the reward-based sampling mechanisms, which currently rely on predefined rules. If these rules are poorly designed or misaligned with user intent, they may inadvertently introduce bias and lead to undesired model behaviors. Additionally, due to the substantial computational overhead associated with self-consuming reinforcement learning, we exclude it from our current experimental setup.

\bibliography{main}

\newpage
\appendix

\section*{Appendix Arrangement}
The Appendix is organized as follows. 
\squishlist
    \item \textbf{Section~\ref{app:sec-broader}: Broader Impact}
    \item \textbf{Section~\ref{app:sec-related}: Related Work}
    \item \textbf{Section~\ref{app:exp}: Experimental details}: Dataset Descriptions (Section~\ref{app:exp:dataset}), Implementation Details (Section~\ref{app:exp:implementation}), Experimental Results (Section~\ref{app:sec:exp-res}), Parameter Study (Section~\ref{app:sec:exp:para})
    \item \textbf{Section~\ref{app:sec-mitigation}: Mitigation Strategy}
    \item \textbf{Section~\ref{app:sec-sft}: Discussion about Focusing on SFT}
    \item \textbf{Section~\ref{app:sec-diss-cycle}: Discussion about Full Synthetic Data Cycle}
    \item \textbf{Section~\ref{app:sec-dpo}: Additional Experiments using DPO}
    \item \textbf{Section~\ref{app:sec-beyond-self}: Additional Experiments beyond Self-consuming}
    \item \textbf{Section~\ref{app:sec-future}: Future Directions}
\squishend

\section{Broader Impact}
\label{app:sec-broader}
Our study highlights how certain types of bias in self-consuming performative loop change.
While we propose mitigation strategies, deploying such self-consuming training pipelines without careful control may unintentionally reinforce existing societal biases or introduce new ones, especially in sensitive applications such as hiring, healthcare, or education. Moreover, the reward-based sampling mechanisms we explore rely on predefined rules, which, if poorly designed or misaligned with user values, may further skew model behavior. As the use of synthetic data becomes more widespread, it is critical to monitor its downstream effects and establish responsible practices for data generation and model retraining.

\section{Related Work}
\label{app:sec-related}
\textbf{Self-consuming Training Loop }
The self-consuming training loop in generative models has recently gained significant attention~\cite{hataya2023will, alemohammad2024self, alemohammad2023self}. Several studies~\cite{alemohammad2023self, shumailov2024ai, briesch2023large} have evidenced catastrophic degradation of generated data in fully synthetic training loops. The phenomenon often referred to as model collapse in recursive training. \citet{seddik2024bad} demonstrate that model collapse is inevitable when training solely on synthetic data from a statistical perspective. Both \citet{alemohammad2023self} and \citet{bertrand2023stability} observe that the inclusion of fresh, real data can stabilize the self-consuming training loop. A recent theoretical contribution by \citet{dohmatob2024model} provides bounds on performance degradation in regression settings, along with modified scaling laws~\citep{dohmatob2024tale}. Several works~\cite{seddik2024bad, briesch2023large, guo2023curious} focus specifically on LLMs, highlighting their unique vulnerabilities to model collapse. In parallel, \citet{ferbach2024self} offer a theoretical study on the role of data curation in iterative retraining of generative models, showing that curated datasets act as an implicit preference optimization mechanism.

\textbf{Performative Prediction }
In the social world, predictions are an intrinsic part of the system, where they inform decisions, beliefs and even influence outcomes.
Performative prediction \citep{perdomo2020performative} provides a framework for studying social predictive systems in which the data distribution changes in response to the deployment of a model~\citep{chen2023performative}. A growing body of work has investigated the stability and optimality of such performative systems~\citep{hardt2023performative, piliouras2023multi, jin2024addressing}.

\textbf{Echo Chambers and User–Technology Feedback.}
Echo chambers~\cite{terren2021echo} describe self-reinforcing dynamics in online platforms, where interactions between users and platform mechanisms amplify specific viewpoints or behaviors over time. Prior work shows that moderation policies, recommendation systems, and user migration jointly shape platform-specific ideological patterns, as observed across communities on Reddit and between platforms such as Reddit and X~\cite{chandrasekharan2017you, horta2021platform, hickey2025x}. These studies highlight how feedback-driven user–technology–userbase cycles can induce systematic distributional shifts in content and behavior.

\textbf{Bias in Self-consuming World }
Recent work~\cite{wyllie2024fairness, wang2024bias, chen2024would} investigates the emergence and amplification of biases in self-consuming training loops. LLMs may internalize inherent biases present in human-generated text~\cite{wang2024jobfair}, or adopt specific political viewpoints reflected in their training data~\cite{haller2023opiniongpt, rettenberger2025assessing, wang2024bias}. To address this issue, \citet{wang2024bias} also propose several techniques for mitigating bias for self-consuming LLMs.

\textbf{Rejection Sampling } Rejection sampling is widely used to generate high-quality synthetic data for supervised fine-tuning of LLMs~\cite{yuan2023scaling, toshniwal2024openmathinstruct, khaki2024rs}. It filters candidate outputs based on predefined heuristics or reward models. \citet{yuan2023scaling} apply this technique to augment training data, while \citet{tong2024dart} propose difficulty-aware rejection sampling, biasing data collection toward harder queries via two strategies: Uniform and Prop2Diff. \citet{li2025fastmcts} introduce FastMCTS to efficiently sample multi-step reasoning data, and \citet{khaki2024rs} propose RS-DPO, which selects contrastive samples using a reward model.
\citet{ferbach2024self} find that synthetic curation helps the model converge toward high reward regions. Inspired by this, we design a simple yet effective bias mitigation strategy that combines rule-based rewards with a reweighting scheme to curate high-quality and less biased data. 

\section{Experimental details}
\label{app:exp}

\subsection{Dataset Descriptions}
\label{app:exp:dataset}
\paragraph{News Datasets} The source dataset is Webis-Bias-Flipper-18~\citep{chen2018learning}. It contains 6458 news articles covering 2781 distinct events from 2012 to 2018. For each article, the dataset provides the title, summary, full content, and an associated political bias label. The bias labels include “From the left” for articles published by left-leaning media outlets, “From the right” for those from right-leaning sources, and “Neutral” for politically unbiased articles. We select all the left-leaning and right-leaning news articles labeled as “From the left” and “From the right,” respectively. Each original news article is tokenized into 256-token blocks, and we select two adjacent blocks as the prompt and the corresponding completion. These prompt–completion pairs are used for both training the political bias classification metric and for the self-performative consuming training loop. Table~\ref{tab:dataset} shows the number of samples for each dataset used during the self performative consuming loop.

\paragraph{Preference Datasets} 
In this task, we utilize three datasets:
(1) \textbf{Databricks-Dolly-15k} is a corpus of over 15,000 instruction-response pairs created by thousands of Databricks employees to enable LLMs to exhibit the interactive behavior characteristic of ChatGPT.
(2) \textbf{CreativeWriting-ShareGPT} is a dataset consisting of prompts designed to elicit creative storytelling, along with the corresponding responses generated by ChatGPT.
(3) \textbf{Preference Dissection} is a benchmark dataset designed to quantify LLM preferences across ten scenarios. It enables decomposition of overall preferences into multiple well-defined attributes, including creative writing and knowledge-aware (non-creative) tasks.

We use the creative writing scenario as the disadvantaged group, and the knowledge-aware (i.e., non-creative) scenario as the advantaged group. Importantly, we only use the Preference Dissection dataset as the test dataset. Accordingly, Dolly is used as the training dataset for non-creative attributes, while CreativeWriting-ShareGPT is used to represent creative attributes in training.

\paragraph{Math Datasets} The source dataset is NuminaMath\citep{li2024numinamath}, a comprehensive collection of 860,000 pairs of competition-level math problems and their solutions. For our study, we categorize problems from Chinese K-12 exams as the easy group and those from international math olympiads as the hard group. Detailed sample statistics are provided in Table~\ref{tab:dataset}. We ensure that the test set is disjoint from the training and validation sets to prevent data leakage.

\paragraph{Clarifying the Rationale for Our Task Choices} Our goal is to study bias dynamics under SCPL, which requires tasks where (1) group structure is explicit, should lead to the “performative”, (2) synthetic data can be generated reliably across generations, and (3) bias effects can be measured continuously over iterative fine-tuning. The three tasks we selected were chosen because they satisfy these criteria:

\textbf{Clear group structure.} Each task provides natural, well-defined group splits (e.g., left/right political leaning, creative/non-creative preferences, easy/hard math problems), enabling controlled bias evaluation.

\textbf{Suitability for multi-generation synthetic data.} Many standard bias benchmarks (e.g., BBQ) are closed-ended, making them unsuitable for recursive data generation. Our tasks support repeated prompting and regeneration across iterations.

\textbf{Grounding in prior work.} Political bias in news generation has been studied~\citep{wang2024bias}; creative vs. non-creative preference gaps follow the Preference Dissection; and easy/hard math disparities capture reasoning performance differences. We will clarify this rationale in Section 4 of the revised version.

\begin{table*}
  \centering
  \begin{tabular}{c|ccc|cc|cc}
    \toprule
    \textbf{Dataset} & \multicolumn{3}{c}{\textbf{Initial Real $D_0$}} &  \multicolumn{2}{c}{\textbf{Candidate $D_{\text{candidate}}$}} & \multicolumn{2}{c}{\textbf{Held out $D_{\text{test}}$}}\\
    \cmidrule(rr){2-4} \cmidrule(rr){5-6}  \cmidrule(rr){7-8} 
    \textbf{Task} & Fixed Size &  $D_a$ & $D_d$ & $D_{\text{candidate}}^a$ & $D_{\text{candidate}}^d$ & $D_{\text{test}}^a$ & $D_{\text{test}}^d$ \\
    \midrule
    News Continuation  & 5000 & 3000 & 2000 & 33873 & 41334 & 500 & 500  \\
    Preference Dissection & 2000 & 1600 & 400 & 13407 & 6253 & - & - \\
    Numina Math & 5000 & 3000 & 2000 & 20000 & 15000 & 500 & 500  \\
    \bottomrule
  \end{tabular}
  \caption{Dataset statistics for the three tasks, showing the number of samples in each dataset at the initial generation 0.}
  \label{tab:dataset}
\end{table*}

\subsection{Implementation Details}
\label{app:exp:implementation}
All experiments can be conducted using two 80 GB A100 GPUs. 

\begin{table*}[t]
\centering
\resizebox{0.9\textwidth}{!}{
\begin{tabular}{llccc}
\hline
\textbf{Setting} & \textbf{Data Type} & \textbf{Group Ratio Change} & \textbf{Keep Accumulated Data} & \textbf{Reuse Prompt} \\
\hline
\multicolumn{5}{c}{\textit{Traditional self-consuming loop (non-dynamic)}} \\
\hline
Traditional (real)       & real         & \xmark & \xmark & \xmark \\
Traditional (syn)        & syn          & \xmark & \xmark & \cmark \\
\hline
\multicolumn{5}{c}{\textit{Proposed self-consuming performative loop (dynamic)}} \\
\hline
real-dynamic             & real         & \cmark & \xmark & \xmark \\
syn-dynamic              & syn          & \cmark & \xmark & \xmark \\
syn-dynamic-fr           & syn          & \xmark & \xmark & \xmark \\
real-dynamic-accu        & real         & \cmark & \cmark & \xmark \\
syn-dynamic-accu         & syn          & \cmark & \cmark & \xmark \\
\hline
\end{tabular}
}
\caption{Comparison of different self-consuming training settings. Dynamic methods use group ratio change, while only the traditional synthetic setting reuses prompts.}
\label{tab:setting_comparison}
\end{table*}

\paragraph{Generation parameters } We used the vLLM~\cite{kwon2023efficient} framework to generate the synthetic data. The generation parameters in the main observation experiments are set as follows: n = 1 (single sample per prompt). All other parameters followed the default settings in vLLM, including temperature = 1.0 and $top_p$ = 1.0. These settings correspond to standard sampling without additional constraints.

\paragraph{Right Lean Bias Evaluation} 
Using an unbiased dataset of 1000 human-written articles evenly distributed across right-lean and left-lean categories, we generate the synthetic articles by deterministically predicting the next 256 tokens. Right lean bias is the percentage of the predicted right lean chunks in the unbiased dataset. If the model had no bias, the bias score should be around 0.5. Following~\citet{wang2024bias}, we train a binary classifier to determine the political leaning of synthetic generation (1 for right-lean, 0 for left-lean). We fine-tuned on Roberta-base model with a learning rate of 2e-5, a batch\_size of 16, weight decay of 0.01 and 5 training epochs. We select the best model based on F1 score (0.94) on the prompts from the unbiased test dataset to have better classification accuracy. 

\paragraph{News Continuation} We use an initial disadvantage ratio $r_d^0$ of 0.4, and set the dataset size for each iteration to 5,000 samples ($n=5,000$). For training, we adopt a learning rate of $2e-5$ and a weight decay of 0.01. Each iteration is trained for 5 epochs, and the entire process consists of 3 generations.
To simulate increasing bias in human preferences, the disadvantage group ratio is linearly reduced across three generations, starting from 0.4 and decreasing to 0.22.

\paragraph{Preference Dissection} We use an initial disadvantage ratio $r_d^0$ of 0.2, and set the dataset size for each iteration to 2,000 samples ($n=2,000$). 
For training, we adopt a learning rate of $2e-5$ and a weight decay of 0.01. Each iteration is trained for 5 epochs, and the entire process consists of five generations.
To simulate increasing bias in human preferences, the disadvantage group ratio is linearly reduced across five generations, starting from 0.2 and decreasing to 0.0.

\paragraph{Numina Math}  We use an initial disadvantage ratio $r_d^0$ of 0.4, and set the dataset size for each iteration to 5,000 samples ($n=5,000$). 
For training, we adopt a learning rate of $2e-5$ and a weight decay of 0.01. Each iteration is trained for 3 epochs, and the entire process consists of five generations.
To simulate increasing bias in human preferences, the disadvantage group ratio is linearly reduced across five generations, starting from 0.4 and decreasing to 0.2.

\begin{figure}[t]
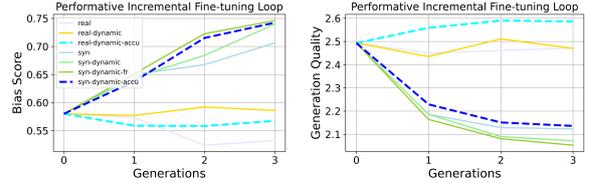

    \centering
    \begin{minipage}[b]{0.49\linewidth}
        \centering
        \includegraphics[width=\linewidth]{section/Figures/news-Llama2-7b-46-5000-finetuning_Bias_Score.png}
    \end{minipage}
    \hfill
    \begin{minipage}[b]{0.49\linewidth}
        \centering
        \includegraphics[width=\linewidth]{section/Figures/news-Llama2-7b-46-5000-finetuning_Generation_Quality.png}
    \end{minipage}
    \caption{The preference bias and generation quality on news task using Llama2-7B in incremental fine-tuning loop.}
    \label{app:fig:news-finetuning}
\end{figure}

\begin{figure}
    \centering
    \begin{minipage}[b]{0.49\linewidth}
        \centering
         \includegraphics[width=\columnwidth]{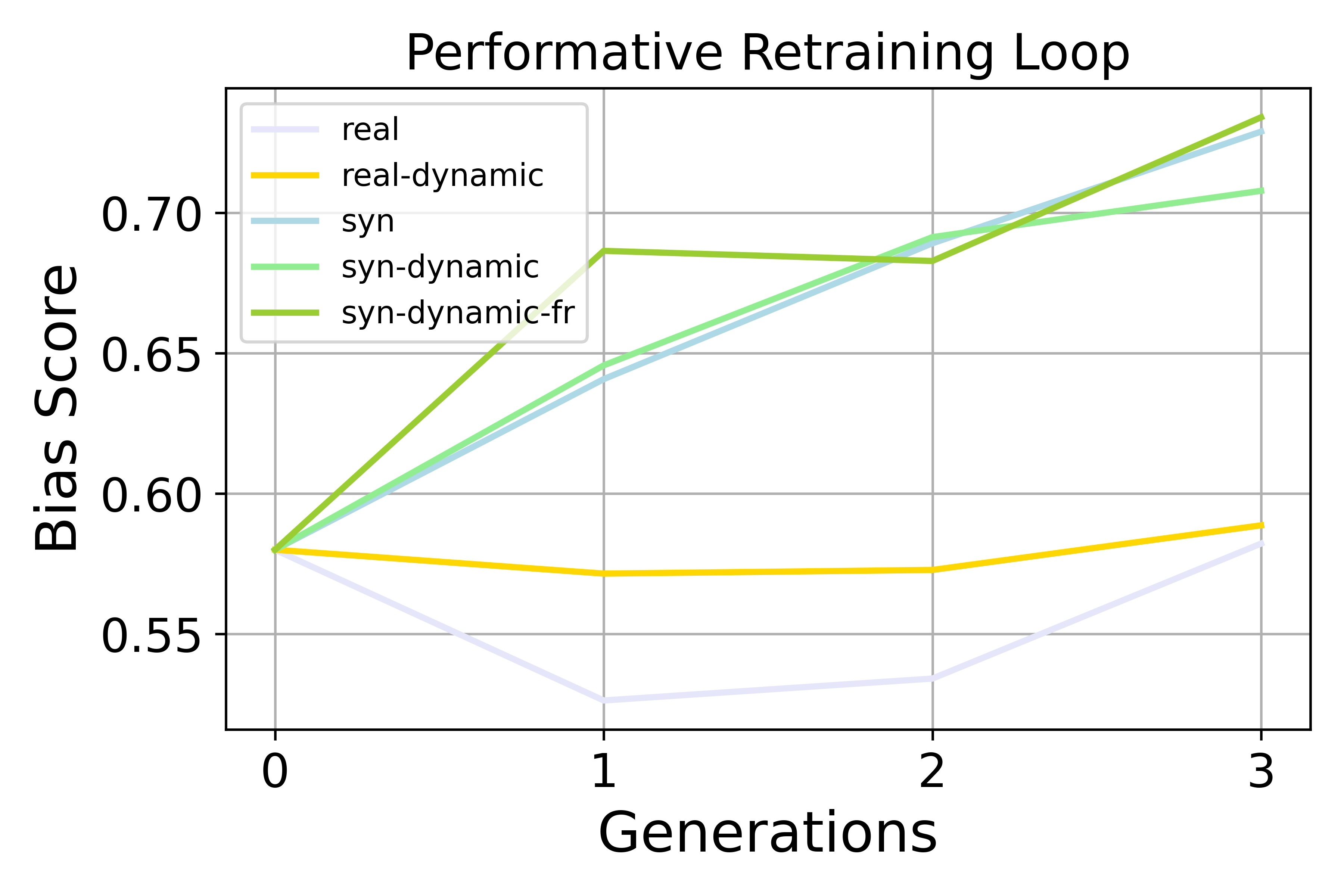}
    \end{minipage}
    \begin{minipage}[b]{0.49\linewidth}
        \centering
          \includegraphics[width=\columnwidth]{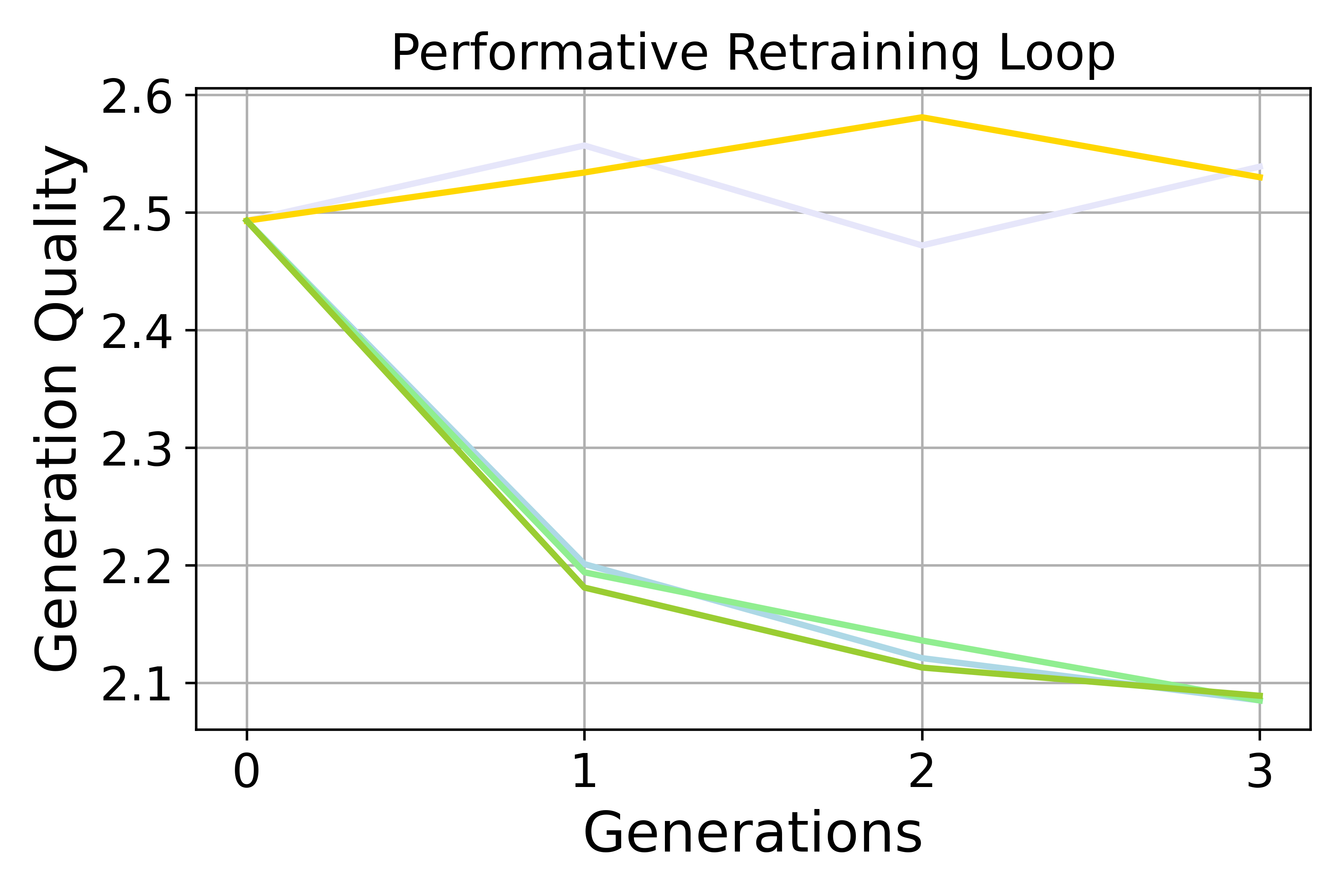}
    \end{minipage}
  \caption{Bias score and generation quality variations in the self-consuming performative retraining loop on the News task using Llama2-7B. }
   \label{app:fig:news-retraining}
\end{figure}

\subsection{Experimental Results}
\label{app:sec:exp-res}
\paragraph{Results in self-consuming performative loop on News task using Llama2-7B~\cite{touvron2023Llama}} Figure~\ref{app:fig:news-finetuning} and Figure~\ref{app:fig:news-retraining} show the variation in bias scores and generation quality in the self-consuming performative fine-tuning loop and retraining loop, respectively. 
The bias score increases more rapidly in the dynamic setting, and the generation quality declines faster under the incremental fine-tuning loop. In contrast, both bias score and generation quality in the performative retraining loop exhibit more stable trends compared with the Non-dynamic training loop. The accumulation of synthetic data fails to mitigate bias in the incremental fine-tuning loop. However, incorporating more accumulated real data improves generation quality throughout the training loop.

\begin{figure*}
    \centering
    \begin{minipage}[b]{0.32\linewidth}
        \centering
         \includegraphics[width=\columnwidth]{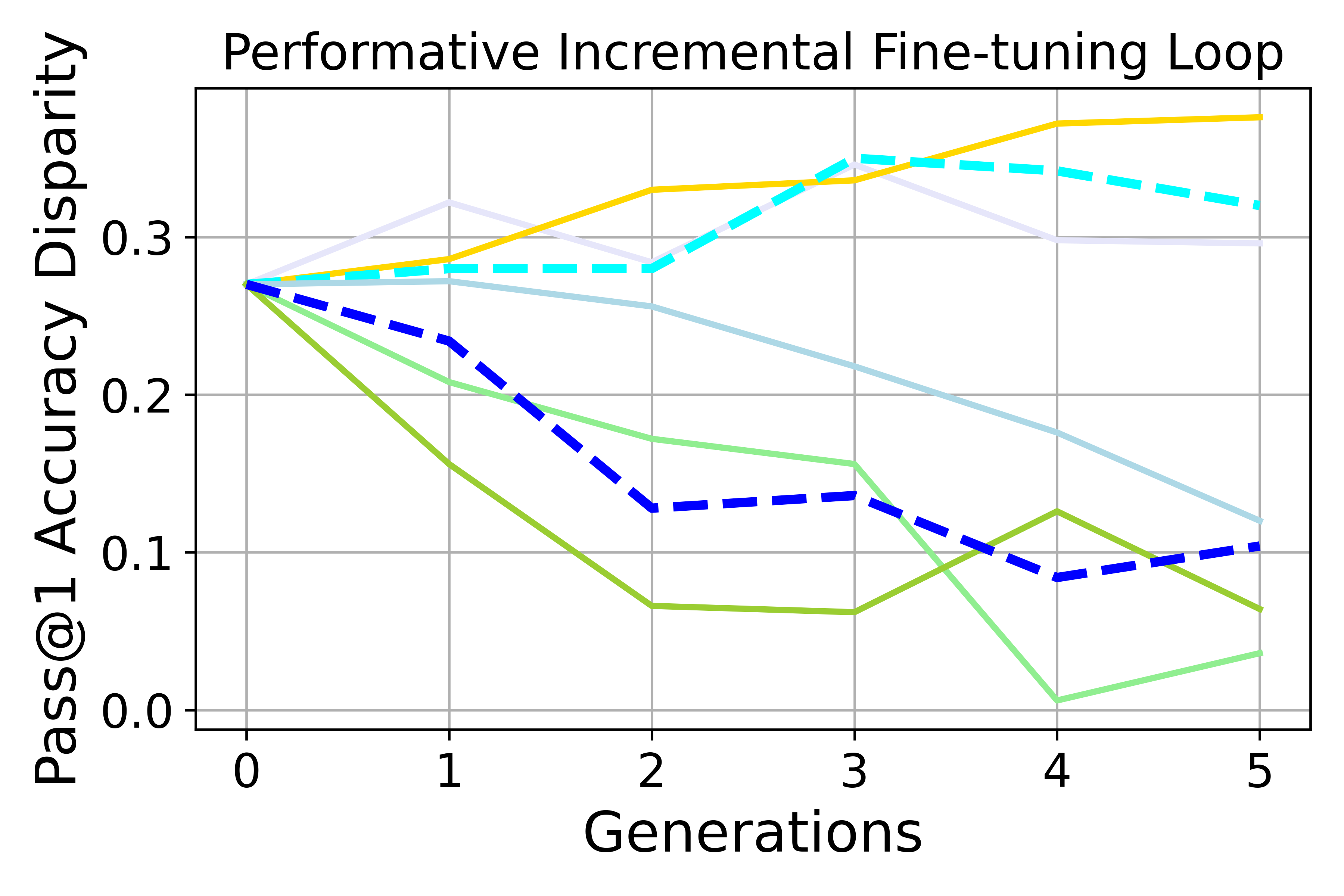}

    \end{minipage}
    \begin{minipage}[b]{0.32\linewidth}
        \centering
         \includegraphics[width=\columnwidth]{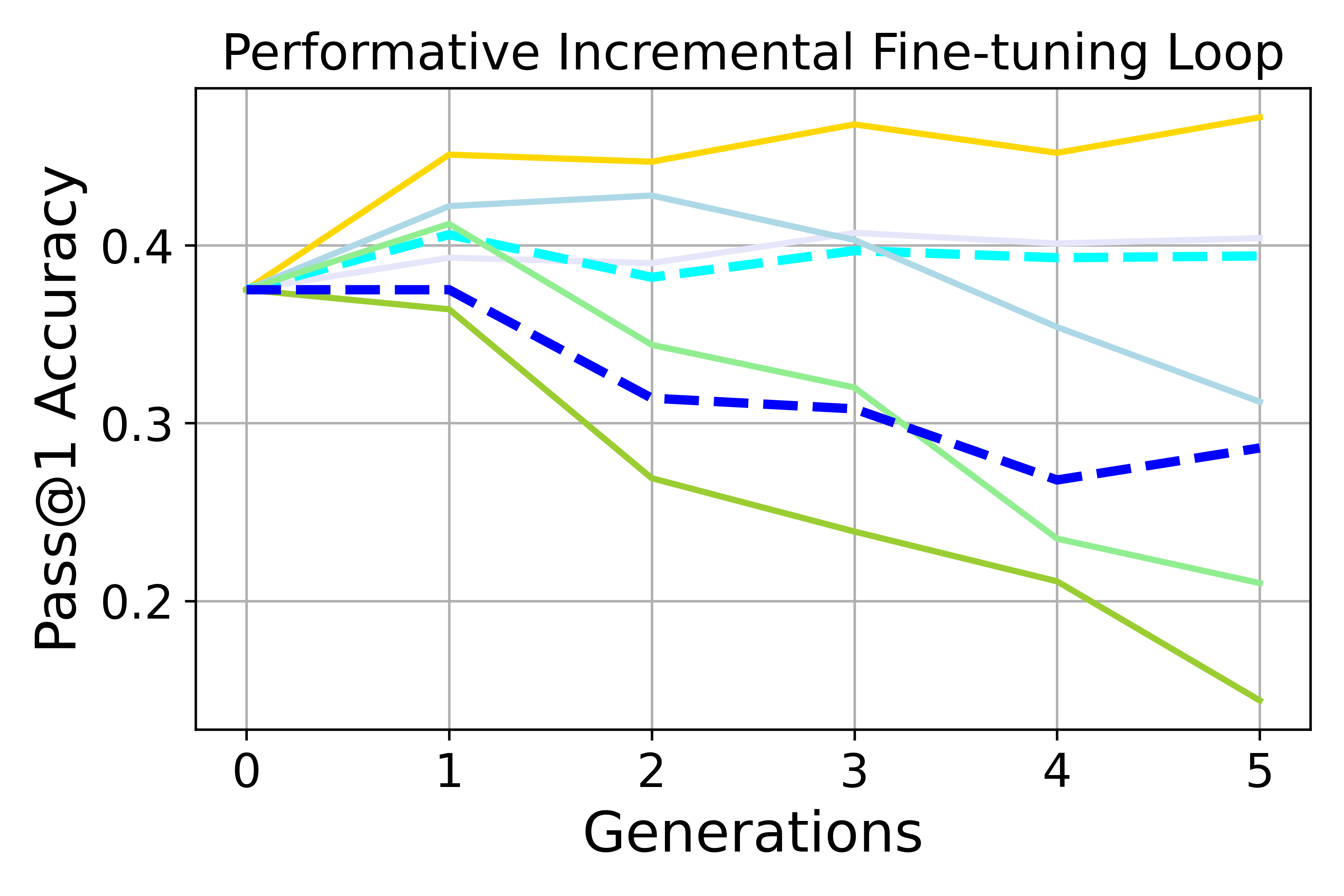}
    \end{minipage}
        \begin{minipage}[b]{0.32\linewidth}
        \centering
         \includegraphics[width=\columnwidth]{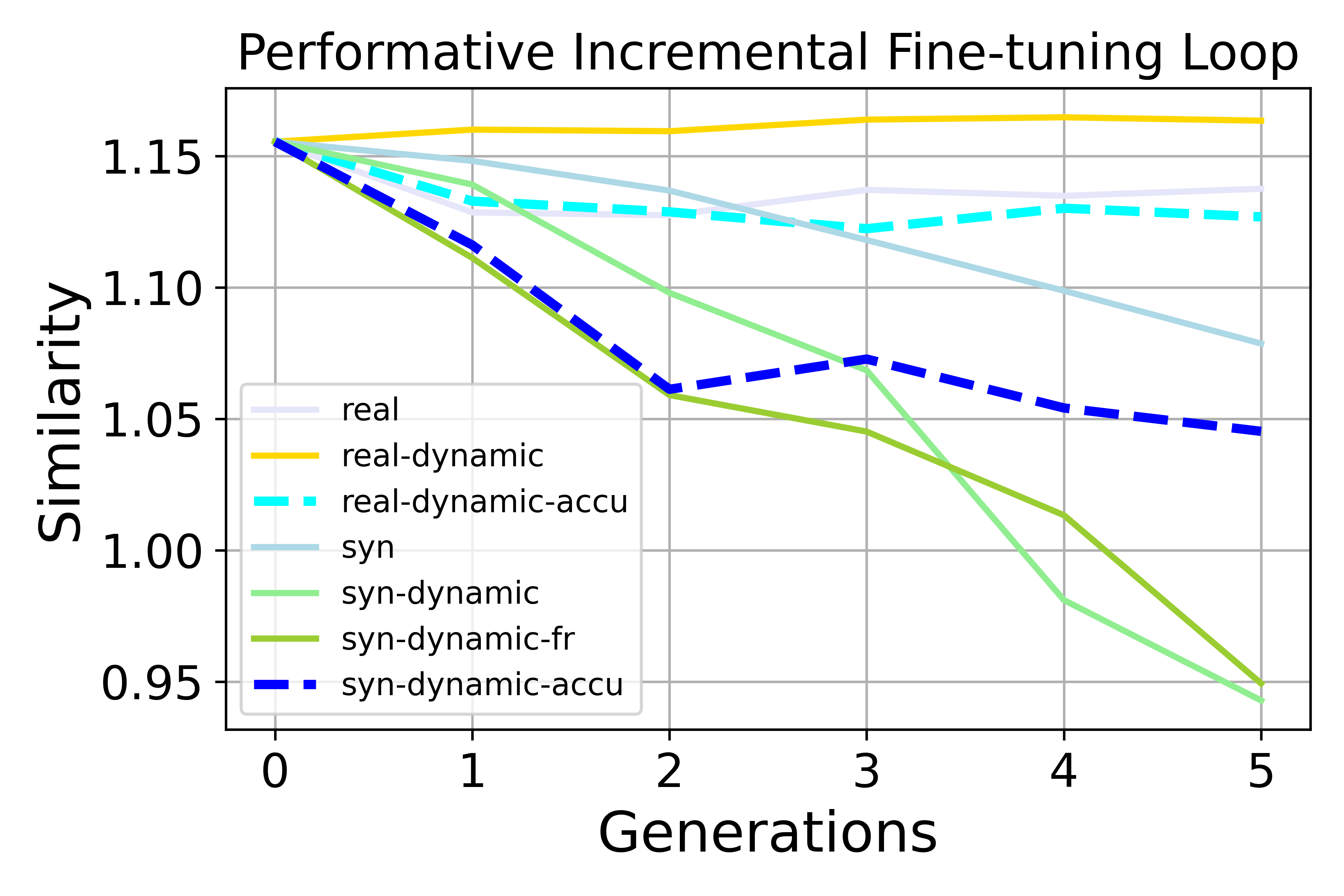}
    \end{minipage}
  \caption{Disparate bias and mathematical problem solving ability variations in the self-consuming performative incremental fine-tuning loop on the Math task using Qwen2.5-Math-7B. }
  \label{app:fig:math-finetuning}
\end{figure*}

\paragraph{Results in self-consuming performative loop on Math task using Qwen2.5-Math-7B~\cite{yang2024qwen2}.} As shown in Figure~\ref{app:fig:math-finetuning}, we observe an interesting trend: while the dynamic setting slows down the performance and disparate bias drop on math-solving tasks for Qwen2.5-Math-1.5B, it leads to faster degradation for Qwen2.5-Math-7B. This may be because larger models are more sensitive to shifts in training data distribution. The dynamic performative training process, while helpful for smaller models by reinforcing useful signals, may disturb the learned balance in larger models and reduce exposure to harder or more diverse samples, thus accelerating performance decline.

\paragraph{Results in the self-consuming performative incremental fine-tuning loop on the News task using Qwen2.5-1.5B, evaluated on generated datasets.}
To better understand the cause of preference bias amplification, we evaluate model performance on the generated datasets (e.g., $D_1$, $D_2$, etc.). As shown in Table~\ref{app:tab:eval}, these datasets inherit existing bias, which is further amplified during the incremental fine-tuning process.

\begin{table*}
  \centering
  \begin{tabular}{lcccc}
    \toprule
    \textbf{Method} & \textbf{$D_0$} &  
    \textbf{$D_1$} & \textbf{$D_2$} &   \textbf{$D_3$}\\
    \midrule
    Synthetic-Non-dynamic & 0.0216 & 0.2240 & 0.3938 & 0.5307 \\
    Synthetic-Dynamic & 0.0216 &0.2158 & 0.4183 & 0.6037 \\
    Synthetic-Dynamic-Fixed Ratio & 0.0216 & 0.2045 & 0.4266 & 0.6143 \\
    \bottomrule
  \end{tabular}
  \caption{Evaluation on generated datasets ($D_1$, $D_2$, ...) shows that preference bias is inherited and amplified during the self-consuming performative incremental fine-tuning loop using Qwen2.5-1.5B.}
  \label{app:tab:eval}
\end{table*}

\begin{table*}
  \centering
  \begin{tabular}{lccccc}
    \hline
    \textbf{Method} & \textbf{Initial Ratio}& \textbf{$\mathcal{M}_0$} &  
    \textbf{$\mathcal{M}_1$} & \textbf{$\mathcal{M}_2$} &   \textbf{$\mathcal{M}_3$}\\
    \hline
    Synthetic-Non-dynamic & 0.4 & 0.5800 & 0.6500 & 0.6675 & 0.7063  \\
    Synthetic-Dynamic & 0.4 &0.5800 & 0.6447 & 0.6839 & 0.7399 \\
    Synthetic-Dynamic-Fixed Ratio & 0.4 & 0.5800 & 0.6512 & 0.7226 & 0.7461 \\
    \hline
     Synthetic-Non-dynamic & 0.2 & 0.6417 & 0.6553 & 0.6623 & 0.6700  \\
    Synthetic-Dynamic & 0.2 & 0.6417 &0.6766 & 0.6554 & 0.6931 \\
    Synthetic-Dynamic-Fixed Ratio & 0.2 & 0.6417 & 0.6736 & 0.6824 & 0.6782  \\
    \hline
  \end{tabular}
  \caption{Parameter study on the impact of the initial disadvantage ratio in the News task using Llama2-7B.}
  \label{app:tab:param-ratio}
\end{table*}

\begin{table*}[t]
\centering
\label{tab:dpo_self_consuming}
\resizebox{\textwidth}{!}{
\begin{tabular}{l|cccc|cccc}
\hline
 & \multicolumn{4}{c|}{\textbf{Preference Bias}} & \multicolumn{4}{c}{\textbf{Generation Quality}} \\
\hline
\textbf{Method} & $\textbf{$\mathcal{M}_0$}$ & $\textbf{$\mathcal{M}_1$}$ & $\textbf{$\mathcal{M}_2$}$ & $\textbf{$\mathcal{M}_3$}$ & $\textbf{$\mathcal{M}_0$}$ & $\textbf{$\mathcal{M}_1$}$ & $\textbf{$\mathcal{M}_2$}$ & $\textbf{$\mathcal{M}_3$}$  \\
\hline
Syn (no-dynamic) & 0.6307 & 0.6606 & 0.6756 & 0.6456 & 2.282 & 2.164 & 2.294 & 2.564 \\
Syn-dynamic in incremental finetuning loop & 0.6307 & 0.6309 & 0.6311 & 0.6129 & 2.282 & 2.136 & 2.244 & 2.497 \\
Syn-dynamic in retraining loop& 0.6307 & 0.6391 & 0.6287 & 0.6247 & 2.282 & 2.242 & 2.442 & 2.436 \\
\hline
\end{tabular}
}
\caption{Results using DPO in self-consuming performative loop on the News continuation task.}
\label{app:tab-dpo}
\end{table*}

\subsection{Parameter Study}
\label{app:sec:exp:para}
\paragraph{The impact of initial disadvantage ratio in News task using Llama2-7B.} Table~\ref{app:tab:param-ratio} presents the study of the parameters on the impact of the initial disadvantage ratio. 
Interestingly, we observe that when the initial disadvantage ratio is lower (e.g., 0.2), the increase in bias over time is less obvious compared to the higher initial ratio (e.g., 0.4). This counter-intuitive result may be due to the fact that models starting from a more imbalanced setting are already biased toward the majority group, making subsequent performative updates reinforce existing preferences rather than introduce significant new bias. In contrast, when the initial distribution is more balanced, the performative feedback loop has a greater opportunity to shift model behavior and amplify bias over time.

\section{Mitigation Strategy}
\label{app:sec-mitigation}
\subsection{Rejection Sampling Baseline}
\label{app:exp-baseline}
\paragraph{Vanilla Rejection Sampling (VRS)}
For each prompt, we generate k candidate responses. For the k generated responses, we evaluate the correctness or the ability to satisfy the predefined standard. 
For news continuation task, the predefined standard include bias, generation quality, similarity to the ground truth continuation, and sentiment score. For bias, we use the trained classifier to determine the score. For generation quality, based on the Gibberish detector, if the Gibberish score is 1 or 0, the score will be -1, otherwise the score will be 1. For similarity, the threshold is 0.5. Then we sum all four scores, if the sum greater than 0, then we think it satisfy the standard. 
After that, we randomly select from these candidates. We make sure at least we will select one response per prompt. 

\paragraph{Top Per Prompt (TPP)} For each prompt, the method generates k responses and selects the response with the highest rewards.

\paragraph{Top Overall Prompts (TOP)} For each prompt, the method generates $k$ responses and selects $n$ response from the $n*k$ candidates with the highest rewards.

\subsection{Reward-based Sampling with Reweight Scheme}

Algorithm~\ref{alg:mitigation} describes the Reward-based Sampling with Reweighting scheme, where $X_{\text{candidate}, t-1}^d$ denotes the set of prompts from the disadvantaged group generated during iteration $t-1$ of the performative loop.

\begin{figure*}
    \centering
    \begin{minipage}[b]{0.45\linewidth}
        \centering
         \includegraphics[width=\columnwidth]{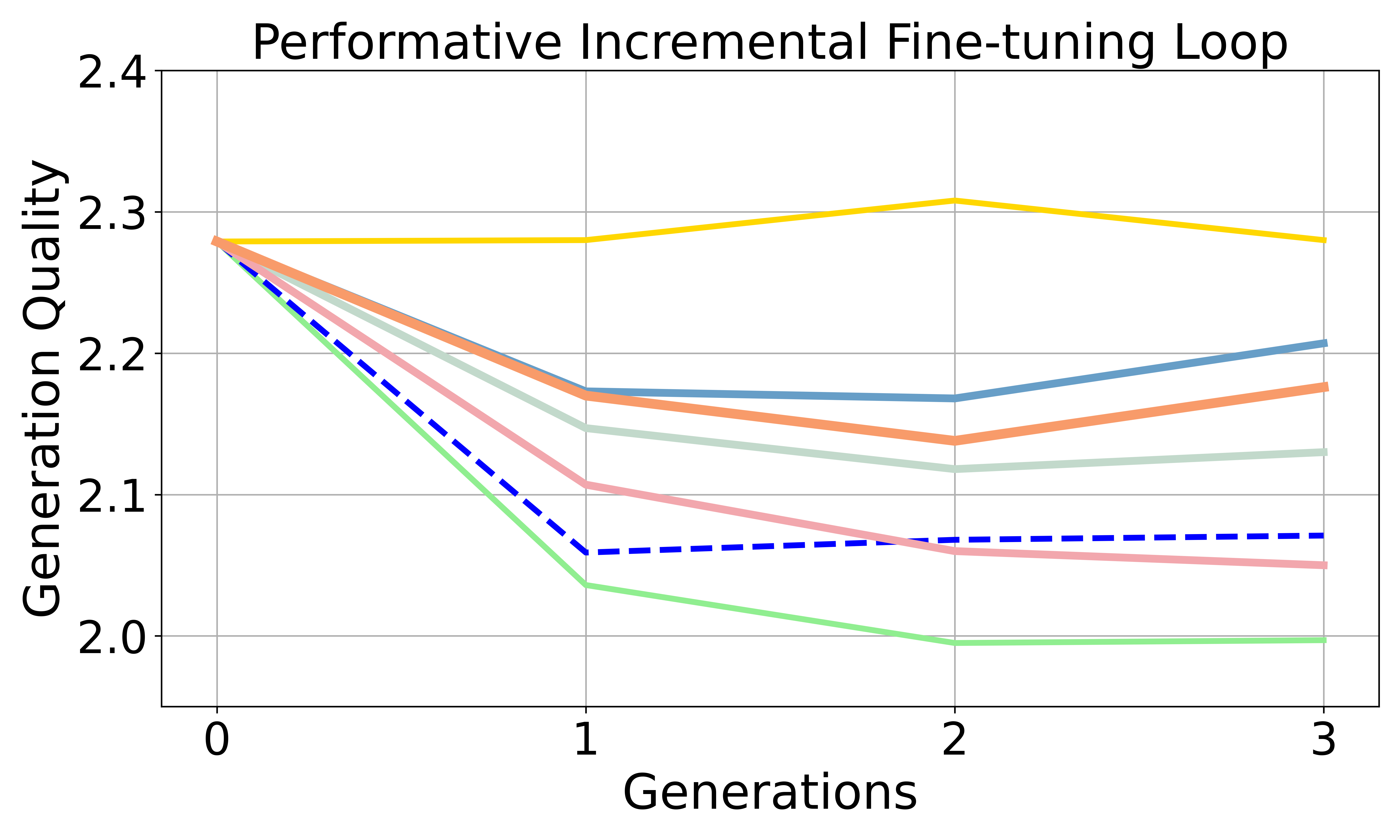}
    \end{minipage}
    \begin{minipage}[b]{0.45\linewidth}
        \centering
          \includegraphics[width=\columnwidth]{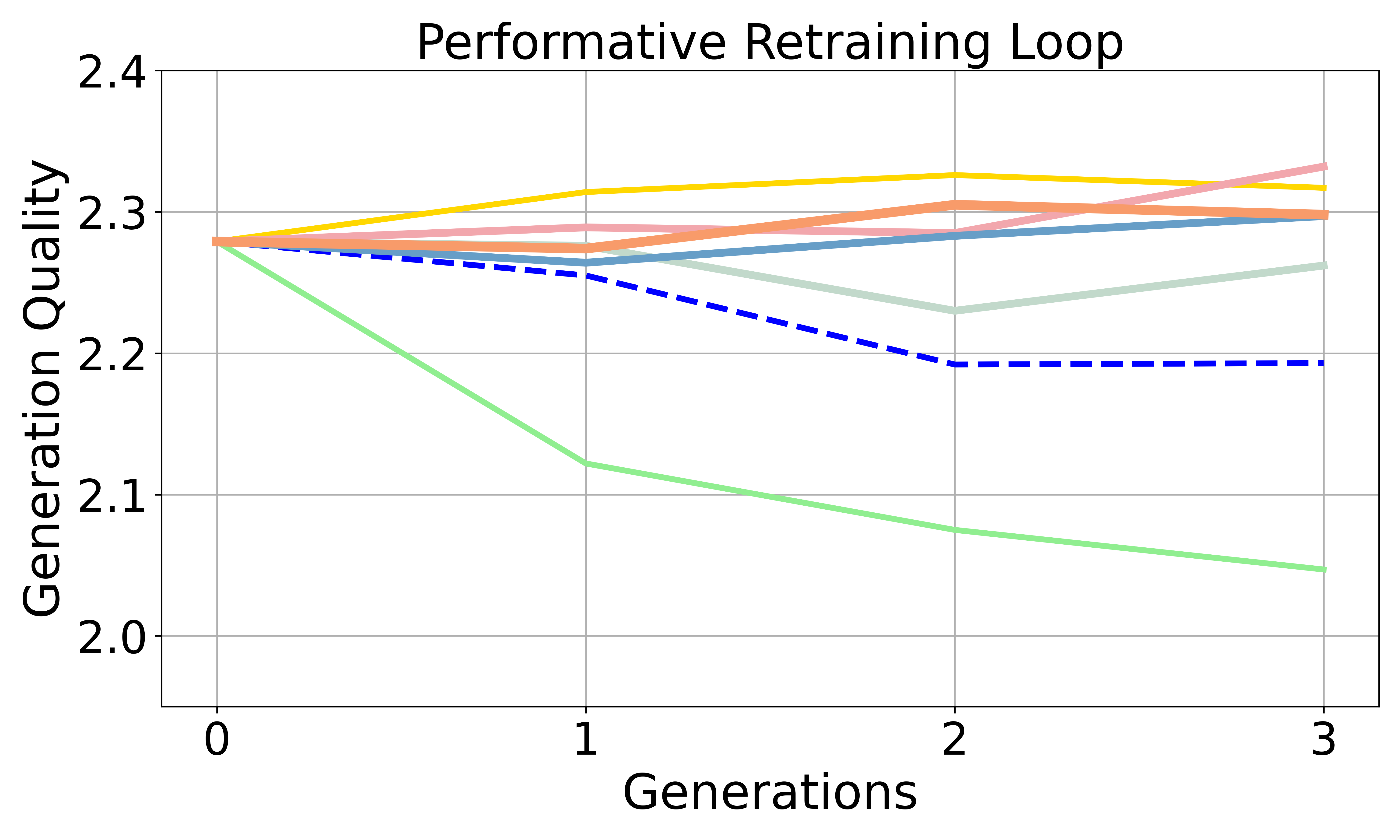}
    \end{minipage}
  \caption{Generation quality variations in the self-consuming performative loop on the News task under different mitigation methods. }
   \label{app:fig:news-mitigation-generation}
\end{figure*}

\begin{algorithm}[tb]
   \caption{Reward-based Reweighting Sampling}
   \label{alg:mitigation}
\begin{algorithmic}
   \State {\bfseries Input:} LLM $\mathcal{M}_{t-1}$, disadvantage ratio $r_d$, $D_{t-1}$, $X_{\text{candidate,t-1}}^a$, $X_{\text{candidate,t-1}}^d$, sample number k
    \State \textbf{Initialize:} $L_a=0$, $L_d=0$, $D_a=[ ]$, $D_d=[ ]$, $D_t = [ ]$, $D_{temp}=[]$, $D_{r}=[]$

    \State $L_a =\frac{1}{4}* |X_{\text{candidate, t-1}}^a| + |X_{\text{candidate, t-1}}^d|$
    \State $L_d = |D_{t-1}|-L_a$
    \State $k_d = L_d // |X_{\text{candidate, t-1}}^d| + 1 + k$ ; $k_a = k $ 
    \For{$x$ {\bfseries in} prompts $X_{\text{candidate, t-1}}^a$}
        \State $\bar{y}_1, \ldots, \bar{y}_{k_a} \sim \mathcal{M}_{t-1}(x)$; 
        \State Calculate reward $R(x,\bar{y})$ for each $(x,\bar{y})$ pair
        \State Add the best pair $(x,y)$ to $D_a$
   \EndFor
   \State Randomly select $L_a$ samples and add to $D_t$ 
    \For{$x$ {\bfseries in} prompts $X_{\text{candidate}}^d$}
        \State $\bar{y}_1, \ldots, \bar{y}_{k_d} \sim \mathcal{M}_{t-1}(x)$; 
        \State Calculate reward $R(x,\bar{y})$ for each $(x,\bar{y})$ pair
        \State Add all pair in $D_{r}$
   \EndFor
\Repeat
     \State For each query $x$ in $D_{r}$, add the best $(x,\bar{y})$ based on $R(x,\bar{y})$ to $D_{temp}$ 
    \State Remove the added pair from $D_r$
\Until{$L_d - |D_{temp}|/|X_{\text{candidate, t-1}}^d|$ is zero }
    \State Randomly select $L_d-|D_{temp}|$ samples and add to $D_t$; Add $D_{temp}$ to $D_t$ 
    \State {\bfseries Return:} $D_t$
\end{algorithmic}
\end{algorithm}

\section{Discussion about Focusing on SFT}
\label{app:sec-sft}
Our current study intentionally focuses on supervised fine-tuning (SFT) to isolate and better understand the dynamics of bias variations under the proposed novel self-consuming performative loops. SFT remains a widely adopted method for improving LLM performance on specific downstream tasks~\cite{ouyang2022training}, especially in practical deployment pipelines where collecting high-quality preference labels for RL is costly or infeasible.

One of our contributions lies in analyzing bias variation within a novel framework of self-consuming performative retraining and an incremental fine-tuning loop for the first time, which models how real-world LLMs may be continuously updated using their own outputs. Furthermore, several prior works~\cite{briesch2023large, seddik2024bad, kazdan2024collapse} have studied similar iterative retraining setups using SFT, but did not explore the bias behavior. Considering this line of research, we emphasize more on the performative feedback loop as a distinct and increasingly relevant phenomenon in real-world deployments.

While we acknowledge the value of RL-based methods, they typically rely on curated high quality data or pairwise comparisons at each iteration, which introduces additional complexities such as label noise, instability, and selection bias. Moreover, methods like RLHF is often resource-intensive and less convenient for continuous training. To keep our scope focused and interpretable, we defer the exploration of RL-based or preference-based alignment strategies to future work.

Most methods do not rely solely on RL but rather use SFT as a foundational step before applying RL or after, as seen in~\citet{guo2025deepseek, team2024qwen2}. Thus, understanding bias behavior under SFT is both fundamental and necessary for guiding future research from the community.
We believe our current SFT-based formulation lays a solid foundation for understanding the core performative feedback dynamics, and we are actively planning to incorporate DPO and ORPO into this framework in subsequent research.

\section{Discussion about Full Synthetic Data Cycle}
\label{app:sec-diss-cycle}
\textbf{Our self-consuming performative loop setting is designed to model a more controlled but increasingly prevalent real-world scenario, where a single model provider collects, fine-tunes, and reuses outputs generated by its own models. }This setup is realistic in many real-world applications. For example, commercial AI service providers often collect user prompts and model generations in production environments (e.g., internal confidential report generation, customer interaction, or proprietary toolchains), then fine-tune the model iteratively based on this data.  When a model performs particularly well for a demographic group (e.g., Group A), its outputs may attract more queries from that group, leading to natural performative feedback, a process we explicitly aim to capture (Figure~\ref{fig:framework} in the main paper). Here, the outputs are not openly mixed with external content but rather used and fine-tuned within a closed feedback loop. In such settings, the provider has full control and visibility over both the model and the data it consumes, making the self-consuming assumption realistic and operationally relevant.

\textbf{Our framework serves as a building block for understanding more complex mixed-model scenarios, and incorporating data generated by other LLMs will introduce more uncontrolled variation.} Our goal is to first establish a clean and analyzable framework in a controlled case for studying how bias varies within the novel self-consuming performative loops, which is a setting that has not been systematically explored in prior work but is highly relevant for real-world, task-specific deployments. Incorporating data generated by other LLMs would introduce uncontrolled variation, making it more difficult to attribute observed bias shifts to specific model behaviors or data sources. We therefore believe that focusing on this controlled full synthetic data case is a necessary and foundational step before extending to more complex, mixed-model generation loops, which serve as an important next step for the research community.

\begin{table*}[t]
\centering
\label{tab:synthetic_loop}
\resizebox{0.9\textwidth}{!}{
\begin{tabular}{ll|cccc|cccc}
\hline
\multicolumn{2}{c|}{} & \multicolumn{4}{c|}{\textbf{Preference Bias}} & \multicolumn{4}{c}{\textbf{Generation Quality}} \\
\hline
\textbf{Method} & \textbf{Ratio} & $\textbf{$\mathcal{M}_0$}$ & $\textbf{$\mathcal{M}_1$}$ & $\textbf{$\mathcal{M}_2$}$ & $\textbf{$\mathcal{M}_3$}$ & $\textbf{$\mathcal{M}_0$}$ & $\textbf{$\mathcal{M}_1$}$ & $\textbf{$\mathcal{M}_2$}$ & $\textbf{$\mathcal{M}_3$}$  \\
\hline
Incremental finetuning & 0.5 & 0.6163 & 0.6802 & 0.6921 & 0.7176 & 2.291 & 2.083 & 2.065 & 2.064 \\
Retraining             & 0.5 & 0.6163 & 0.6847 & 0.6899 & 0.6956 & 2.294 & 2.119 & 2.081 & 2.096 \\
\hline
Incremental finetuning & 0.2 & 0.6048 & 0.6987 & 0.6897 & 0.6737 & 2.282 & 2.068 & 2.102 & 2.092 \\
Retraining             & 0.2 & 0.6048 & 0.6802 & 0.6871 & 0.6817 & 2.282 & 2.117 & 2.135 & 2.124 \\
\hline
\end{tabular}
}
\caption{Results under the synthetic performative loop. The ratio indicates the percentage of self-consuming synthetic data used during training. Ratio denotes the percentage of synthetic data generated from another LLM.}
\label{app:tab-beyond-self}
\end{table*}

\section{Additional Experiments using DPO}
\label{app:sec-dpo}
We conduct preliminary experiments using DPO on the News Continuation task to investigate how bias evolves during iterative preference-based training. In our setup, we use generation quality as the sole scoring metric to construct preference pairs, deliberately excluding any explicit bias control mechanisms.

From Table~\ref{app:tab-dpo}, we observe that bias does not increase in the self-consuming performative loop using DPO. It shows a slight decrease from generation 2 to generation 3. This opens up a promising direction for understanding and addressing bias in iterative preference-based training. In terms of generation quality, we see a clear improvement. This indicates that in a self-consuming iterative preference training loop, using generation quality-guided preference pair construction can meaningfully enhance output quality.

The observed variation are different from those under SFT, here bias-reduction patterns and quality improvements appear under DPO. This is expected because preference-based learning explicitly trains the model to favor higher-quality responses over lower-quality ones. By repeatedly reinforcing distinctions between “good” and “bad” pairs, the model learns clearer decision boundaries. As a result, undesirable or biased patterns become less preferred, while coherent and balanced responses are amplified, naturally leading to both bias reduction and improved overall quality.

\section{Additional Experiments beyond Self-consuming}
\label{app:sec-beyond-self}
The proposed setting serves as a foundational framework for analyzing bias variation, which can be extended to more complex settings involving multiple models and data sources. We conduct additional experiments involving both self-generated data and data generated by another LLM.

To investigate how bias trends change when mixing data from multiple generation sources, we split the real prompts into two groups (1:1 ratio). One group is used to generate data with the self-consuming model (Qwen2.5-1.5B), and the other group is used with an external open-source model (Llama3.1-8B). To ensure fair comparison, we control the dataset size across iterations. At each training step t, we collect: 3000 prompts from advantage group and 2000 from disadvantage group, and generate 1500 + 1000 synthetic preference pairs from Qwen2.5-1.5B, and another 1500 + 1000 pairs from Llama3.1-8B.

We specifically avoid using closed-source APIs to mitigate risks of data leakage as synthetic training data is generated using fresh real prompts collected from users. Future work will explore a wider range of models, settings, and mixing strategies to more fully understand bias propagation in multi-model synthetic feedback loops, which is beyond the current paper’s scope.

In Table~\ref{app:tab-beyond-self}, when incorporating synthetic data from another LLM at a ratio of 0.5, the bias score in the synthetic performative loop still exhibits an upward trend similar to that of the pure self-consuming setup (Figure~\ref{fig:news} in the main paper), albeit with a slower rate of increase. The trend in generation quality remains largely unchanged. When the ratio is further reduced (0.2), the bias score no longer shows an increasing trend, indicating that the proportion of self-generated synthetic data influences the bias dynamics in the performative loop.

\section{Future Directions}
\label{app:sec-future}
In this work, we primarily focus on supervised fine-tuning, while also conducting preliminary experiments with direct preference optimization. Our framework offers a promising direction through dynamic control, which can be further extended in future work to incorporate alternative training strategies such as reinforcement learning with human feedback (RLHF) and preference learning (DPO and SimPO). We believe that our current SFT-based formulation provides a strong foundation for understanding the core dynamics of performative feedback, and that integrating DPO or reinforcement learning methods (e.g., PPO) could further enhance its flexibility and effectiveness. 
As synthetic data can be generated using multiple models, exploring performative bias beyond self-consuming settings presents a promising future direction. However, studying multi-model feedback loops introduces substantial design complexity. Key factors, such as the number and scale of participating models, as well as the proportion of self-generated data, require careful control and systematic evaluation. We believe that addressing these challenges would greatly benefit from coordinated, community-wide efforts.

\end{document}